\documentclass[10pt,journal,compsoc]{IEEEtran}
\ifCLASSOPTIONcompsoc
 \usepackage{cite}
\else
 \usepackage{cite}
\fi
\usepackage{xspace}
\usepackage{tabularx, booktabs}

\ifCLASSINFOpdf
 \usepackage[pdftex]{graphicx}
 \graphicspath{{Images/}}
 \DeclareGraphicsExtensions{.pdf,.jpeg,.png}
\else
\fi
\usepackage{booktabs}

\usepackage{algorithmic}

\usepackage{xcolor}

\ifCLASSOPTIONcompsoc
 \usepackage[caption=false,font=footnotesize,labelfont=sf,textfont=sf]{subfig}
 
\else
 \usepackage[caption=false,font=footnotesize]{subfig}
\fi

\usepackage{color}
\usepackage[algo2e,ruled,vlined]{algorithm2e}
\usepackage{amsmath}
\usepackage{amssymb}
\usepackage{hyperref}
\usepackage{xcolor}
\usepackage[]{algorithm}
\usepackage{comment}
\usepackage{multirow}

\newcommand*{\colorboxed}{}
\def\colorboxed#1#{%
 \colorboxedAux{#1}%
}
\newcommand*{\colorboxedAux}[3]{%
 \begingroup
 \colorlet{cb@saved}{.}%
 \color#1{#2}%
 \boxed{%
 \color{cb@saved}%
 #3%
 }%
 \endgroup
}

\usepackage{array}
\newcolumntype{L}[1]{>{\raggedright\let\newline\\\arraybackslash\hspace{0pt}}m{#1}}
\newcolumntype{C}[1]{>{\centering\let\newline\\\arraybackslash\hspace{0pt}}m{#1}}
\newcolumntype{R}[1]{>{\raggedleft\let\newline\\\arraybackslash\hspace{0pt}}m{#1}}

\newcommand{\beq}{\begin{equation}}
\newcommand{\eeq}{\end{equation}}

\renewcommand{\vec}[1]{\mathbf{#1}}

\newcommand{\mr}[1]{\mathrm{#1}}

\newcommand{\xx}{\vec{x}}

\newcommand{\vt}{\boldsymbol{\theta}}
\newcommand{\vold}[1]{$#1\!\times\!#1\!\times\!#1$}

\newcommand{\HyperDenseNet}{HyperDenseNet\xspace}

\begin{document}
%

\title{HyperDense-Net: A hyper-densely connected CNN for multi-modal image segmentation}
%
%
%
%

\author{Jose~Dolz,
 Karthik Gopinath,
 Jing Yuan,
 Herve Lombaert,
 Christian~Desrosiers,
 and~Ismail~Ben~Ayed
\IEEEcompsocitemizethanks{\IEEEcompsocthanksitem J. Dolz, K. Gopinath, H. Lombaert, C. Desrosiers and I. Ben Ayed are with the \'Ecole de technologie Superieure, Montreal, Canada. email:jose.dolz@livia.etsmtl.ca}

\IEEEcompsocitemizethanks{\IEEEcompsocthanksitem J. Yuan is with the Xidian University, School of Mathematics and Statistics, Xi'an, China.}
\thanks{Manuscript received XXX; revised XXX.}}

\IEEEpubid{\begin{minipage}{\textwidth}\ \\[12pt]
  Copyright \copyright 2018 IEEE. Personal use of this material is permitted.Permission from IEEE must be obtained for all other uses, including reprinting/republishing this material for advertising or promotional purposes, collecting new collected works for resale or redistribution to servers or lists, or reuse of any copyrighted component of this work in other works.
\end{minipage}}



\IEEEtitleabstractindextext{%
\begin{abstract}
Recently, dense connections have attracted substantial attention in computer vision because they facilitate gradient flow and implicit deep supervision during training.
Particularly, DenseNet, which connects each layer to every other layer in a feed-forward fashion, has shown impressive performances in natural image classification tasks.
We propose {\em HyperDenseNet}, a 3D fully convolutional neural network that extends the definition of dense connectivity to multi-modal segmentation problems.
Each imaging modality has a path, and dense connections occur not only between the pairs of layers within the same path, but also between those across different paths. 
This contrasts with the existing multi-modal CNN approaches, in which modeling several modalities relies entirely on a single joint layer (or level of abstraction) for fusion, typically
either at the input or at the output of the network. Therefore, the proposed network has total freedom to learn more complex combinations between the modalities, {\em within and in-between
all the levels of abstraction}, which increases significantly the learning representation. 
We report extensive evaluations over two different and highly competitive multi-modal brain tissue segmentation challenges, iSEG 2017 and MRBrainS 2013, with the former focusing on 6-month infant data and the latter on adult images. {\em HyperDenseNet} yielded significant improvements over many state-of-the-art segmentation networks, ranking at the top on both benchmarks. We further provide a comprehensive experimental analysis of features re-use, which confirms the importance of hyper-dense connections in multi-modal representation learning. Our code is publicly available\footnote{\url{https://www.github.com/josedolz/HyperDenseNet}}.

\end{abstract}

}

\maketitle

\IEEEdisplaynontitleabstractindextext

%
\IEEEpeerreviewmaketitle

\IEEEraisesectionheading{\section{Introduction}\label{sec:introduction}}

%
%
%
%

\begin{figure}[h!]
 \begin{center}
 \mbox{
 \includegraphics[height=0.35\linewidth]{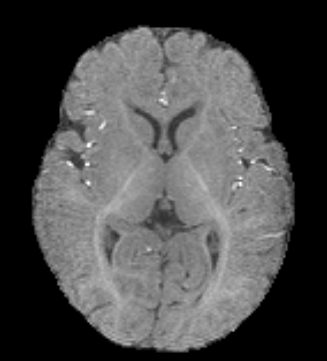}
 \hspace{-2.5 mm}
 \includegraphics[height=0.35\linewidth]{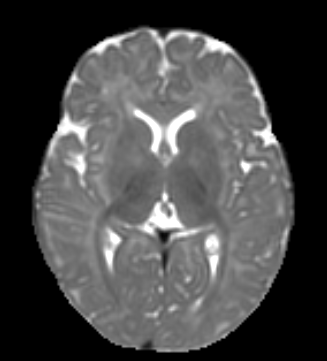}
 \hspace{-2.5 mm}
 \includegraphics[height=0.35\linewidth]{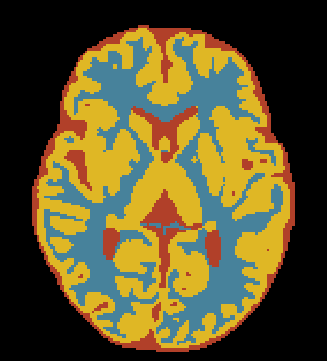}
 }
\caption{Example of data from a training subject. Neonatal isointense brain images from a mid-axial T1 slice (\textit{left}), the corresponding T2 slice (\textit{middle}), and manual segmentation (\textit{right}).}
\label{fig:isointense}
\end{center}
\end{figure}

\IEEEPARstart{M}{ulti-modal} imaging is of primary importance for developing comprehensive models of pathologies and increasing the statistical power of current imaging biomarkers \cite{delbeke2009hybrid}. In neuroimaging studies, different magnetic resonance imaging (MRI) modalities are often combined to overcome the limitations of independent imaging techniques. While T1-weighted images yield a good contrast between gray matter (GM) and white matter (WM) tissues, T2-weighted and proton density (PD) pulses help visualize tissue abnormalities like lesions. Likewise, fluid attenuated inversion recovery (FLAIR) images can enhance the image contrast of white matter lesions resulting from multiple sclerosis \cite{llado2012segmentation}. In brain segmentation, considering multiple MRI modalities is essential to obtain accurate results. This is particularly true for the segmentation of infant brains, where tissue contrast is low (Fig. \ref{fig:isointense}).


Advances in multi-modal imaging, however, come at the price of an inherently large amount of data, imposing a burden on disease assessments. Visual inspections of such an enormous amount of medical images are prohibitively time-consuming, prone to errors and unsuitable for large-scale studies. Therefore, automatic and reliable multi-modal segmentation algorithms are of high interest to the clinical community.

\subsection{Prior work}
\label{ssec:prior}

Multi-modal image segmentation in brain-related applications has received a substantial research attention, for instance, brain tumors \cite{menze2015multimodal,havaei2016hemis,kamnitsas2017efficient,fidon2017scalable}, brain tissues of both infant \cite{prastawa2005automatic,weisenfeld2006segmentation,anbeek2008probabilistic,weisenfeld2009automatic,wang2011automatic,srhoj2012automatic,wang2012atlas,wang2014segmentation,zhang2015deep,nie2016fully,dolz2017Ensemble} and adult \cite{mendrik2015mrbrains,chen2017voxresnet}, subcortical structures \cite{deoni2007segmentation}, among other problems \cite{commowick2016msseg,kamnitsas2017unsupervised,valverde2017improving}. 
Atlas-propagation approaches are commonly used in multi-modal scenarios \cite{gonzalez2016review,makropoulos2017review}. These methods rely on registering one or multiple atlases to the target image, followed by a propagation of manuals labels. When several atlases are considered, labels from individual atlases can be combined into a final segmentation via a label fusion strategy \cite{weisenfeld2006segmentation,weisenfeld2009automatic,wang2012atlas}. When relying solely on atlas fusion, the performance of such techniques might be limited and prone to registration errors. Parametric or deformable models \cite{wang2011automatic} can be used to refine prior estimates of tissue probability \cite{wang2014segmentation}. For example, the study in \cite{wang2014segmentation} investigated a patch-driven method for neonatal brain tissue segmentation, integrating the probability maps of a subject-specific atlas into a level-set framework.

More recently, our community has witnessed a wide adoption of deep learning techniques, particularly, convolutional neural networks (CNNs), as an effective alternative to traditional segmentation approaches. CNN architectures are supervised models, trained end-to-end, to learn a hierarchy of image features representing different levels of abstraction. In contrast to conventional classifiers based on hand-crafted features, CNNs can learn both the features and classifier simultaneously, in a data-driven manner. They achieved state-of-the-art performances in a broad range of medical image segmentation problems \cite{DolzNeuro2017,fechter2017}, including multi-modal tasks \cite{kamnitsas2015multi,zhang2015deep,havaei2016hemis,moeskops2016automatic,nie2016fully,chen2017voxresnet,dolz2017Ensemble,fidon2017scalable,kamnitsas2017efficient,kamnitsas2017unsupervised,valverde2017improving}.

\subsubsection{Fusion of multi-modal CNN feature representations}
 
Most of the existing multi-modal CNN segmentation techniques followed an {\em early-fusion} strategy, which integrates the multi-modality information from the original space of low-level features \cite{zhang2015deep,moeskops2016automatic, kamnitsas2015multi, kamnitsas2017efficient, chen2017voxresnet, valverde2017improving}. For instance, in \cite{zhang2015deep}, MRI T1, T2 and fractional anisotropy (FA) images are simply merged at the input of the network. However, as argued in \cite{Srivastava14} in the context of multi-modal learning, it is difficult to discover highly non-linear relationships between the low-level features of different modalities, more so when such modalities have significantly different statistical properties. In fact, early-fusion methods implicitly assume that the relationship between different modalities are simple (e.g., linear). For instance, the early fusion in \cite{zhang2015deep} learns complementary information from T1, T2 and FA images. However, the relationship between the original T1, T2 and FA image data may be much more complex than complementarity, due to significantly different image acquisition processes \cite{nie2016fully}. The work in \cite{nie2016fully} advocated {\em late fusion} of high-level features as a way that accounts better for the complex relationships between different modalities. They used an independent convolutional network for each modality, and fused the outputs of the different networks in higher-level layers, showing better performance than early fusion in the context infant brain segmentation. These results are in line with a recent study in the machine learning community \cite{Srivastava14}, which 
investigated multimodal learning with deep Boltzmann machines in the context of fusing data from color images and text.




\begin{table*}[t!]
\begin{center}
\begin{scriptsize}
\centering
\renewcommand{\arraystretch}{1.25}
\caption{Overview of representative works on multi-modal brain segmentation.}
\label{table:table_summaryMethods}
\begin{tabular}{lC{30mm}C{30mm}C{39mm}}
\toprule
\textbf{Work} & \textbf{Modality} & \textbf{Target} & \textbf{Method}
 \\
\midrule\midrule

Prastawa et al., 2005~ \cite{prastawa2005automatic} & T1,T2 & Infant brain tissue & Multi-atlas\\
Weisenfeld et al., 2006~ \cite{weisenfeld2006segmentation} & T1,T2 & Infant brain tissue & Multi-atlas\\
Deoni et al., 2007~ \cite{deoni2007segmentation} & T1,T2 & Thalamic nuclei & K-means clustering\\
Anbeek et al., 2008~ \cite{anbeek2008probabilistic} & T2,IR & Infant brain tissue & KNN\\
Weisenfeld and Warfield, 2009~ \cite{weisenfeld2009automatic} & T1,T2 & Infant brain tissue & Multi-atlas\\
Wang et al., 2011~ \cite{wang2011automatic} & T1,T2,FA & Infant brain tissue & Multi-atlas + Level sets\\
Srhoj et al., 2012~ \cite{srhoj2012automatic} & T1,T2 & Infant brain tissue & Multi-atlas + KNN\\
Wang et al., 2012~ \cite{wang2012atlas} & T1,T2 & Infant brain tissue & Multi-atlas\\
Wang et al., 2014~ \cite{wang2014integration} & T1,T2,FA & Infant brain tissue & Multi-atlas + Level sets\\
Kamnitsas et al., 2015~ \cite{kamnitsas2015multi} & Flair, DWI, T1, T2 & Brain lesion & 3D FCNN + CRF\\
Zhang et al., 2015~ \cite{zhang2015deep} & T1,T2,FA & Infant brain tissue & 2D CNN\\
Havaei et al., 2016~ \cite{havaei2016hemis} & T1,T1c,T2,FLAIR & Multiple Sclerosis/Brain tumor & 2D CNN \\
Nie et al., 2016~ \cite{nie2016fully} & T1,T2,FA & Infant brain tissue & 2D FCNN\\
Chen et al., 2017~ \cite{chen2017voxresnet} & T1,T1-IR,FLAIR & Brain tissue & 3D FCNN\\
Dolz et al., 2017~ \cite{dolz2017Ensemble} & T1,T2 & Infant brain tissue & 3D FCNN\\
Fidon et al., 2017~ \cite{fidon2017scalable} & T1,T1c,T2,FLAIR & Brain tumor & CNN \\
Kamnitsas et al., 2017~ \cite{kamnitsas2017efficient} & \begin{tabular}{@{}c@{}}T1,T1c,T2,FLAIR \\ MPRAGE,FLAIR,T2,PD\end{tabular} & Brain tumour/lesions & 3D FCNN + CRF\\ 
Kamnitsas et al., 2017~ \cite{kamnitsas2017unsupervised} &MPRAGE,FLAIR,T2,PD & Traumatic brain injuries & 3D FCNN(Adversarial Training)\\ 
Valverde et al., 2017~ \cite{valverde2017improving} & T1, T2,FLAIR & Multiple-sclerosis & 3D FCNN \\ 
\bottomrule
\end{tabular}
\end{scriptsize}
\end{center}
\end{table*}

\subsubsection{Dense connections in deep networks}

Since the recent introduction of residual learning in \cite{he2016deep}, shortcut connections from early to late layers have become very popular in a breadth of computer vision problems \cite{huang2016deep,larsson2016fractalnet}. Unlike traditional networks, these connections back-propagate gradients directly, thereby mitigating the gradient-vanishing problem and allowing deeper networks. Furthermore, they transform a whole network into a large ensemble of shallower networks, yielding competitive performances in various applications \cite{zagoruyko2016wide,chen2017voxresnet,ranjan2017hyperface,szegedy2017inception}. DenseNet \cite{huang2017densely} extended the concept of shortcut connections, with the input of each layer corresponding to the outputs from all previous layers. Such a dense network facilitates the gradient flow and the learning of more complex patterns, which yielded significant improvements in accuracy and efficiency for natural image classification tasks \cite{huang2017densely}. Inspired by this success, recent works have included dense connections in deep networks for medical image segmentation \cite{li2017h,yu2017automatic,chenmri}. However, these works have either considered a single modality \cite{li2017h,yu2017automatic} or have simply concatenated multiple modalities in a single stream \cite{chenmri}. So far, the impact of dense connectivity across multiple network paths, and its application to multi-modal image segmentation, remains unexplored.


\subsection{Contributions} 



We propose {\em HyperDenseNet}, a 3D fully convolutional neural network that extends the definition of dense connectivity to multi-modal segmentation problems.
Each imaging modality has a path, and dense connections occur not only between the pairs of layers within the same path, but also between those across different paths; see the
illustration in Fig. \ref{fig:net}. This contrasts with the existing multi-modal CNN approaches, in which modeling several modalities relies entirely on a single joint layer (or level of abstraction) for fusion, typically either at the input (early fusion) or at the output (late fusion) of the network. Therefore, the proposed network has total freedom to learn more complex
combinations between the modalities, {\em within and in-between all the levels of abstractions}, which increases significantly the learning representation in comparison to early/late fusion. Furthermore, hyper-dense connections facilitate the learning as they improve gradient flow and impose implicit deep supervision. We report extensive evaluations over two different\footnote{iSEG 2017 focuses on 6-month infant data, whereas MRBrainS 2013 uses adult data. Therefore, there are significant differences between the two benchmarks in term of image data characteristics, e.g, the voxel spacing and number of available modalities.} and highly competitive multi-modal brain tissue segmentation challenges, iSEG 2017 and MRBrainS 2013. {\em HyperDenseNet} yielded significant improvements over many state-of-the-art segmentation networks, ranking at the top on both benchmarks. We further provide a comprehensive experimental analysis of features re-use, which confirms the importance of hyper-dense connections in multi-modal representation learning. Our code is publicly available\footnote{\url{https://www.github.com/josedolz/HyperDenseNet}}.

A preliminary conference version of this work appeared at ISBI 2018 \cite{DolzISBI2018}. This journal version is a substantial extension, including (1) 
a much broader, more informative/rigorous treatment of the subject in the general context of multi-modal segmentation; and (2) comprehensive experiments
with additional baselines and publicly available benchmarks, as well as a thorough investigation of the practical usefulness and impact of hyper-dense connections.

\section{Methods and Materials}

Convolutional neural networks (CNNs) are deep models that can learn feature representations automatically from the training data. They consist of multiple layers, each processing the imaging data at a different level of abstraction, enabling segmentation algorithms to learn from large datasets and discover complex patterns that can be further employed for predicting unseen samples. The first attempts to use CNNs in segmentation problems followed a sliding-window strategy, where the regions defined by the window are processed independently, which impedes segmentation accuracy and computational efficiency. To overcome these limitations, the network can be viewed as a single non-linear convolution, which is trained end-to-end, a process known as fully CNN (FCNN) \cite{FCN}. The latter brings several advantages over standard CNNs. It can handle images of arbitrary sizes and avoid redundant convolution and pooling operations, enabling computationally efficient learning.

\subsection{The proposed Hyper-Dense network}
\label{sec:hyperDense}


The concept of ``\emph{the deeper the better}'' is considered as a key principle in deep learning \cite{he2016deep}. Nevertheless, one obstacle when dealing with deep architectures is the problem of vanishing/exploding gradients, which hampers convergence during training. To address these limitations in very deep architectures, the study in \cite{huang2017densely} investigated densely connected networks. DenseNets are built on the idea that adding direct connections from any layer to all the subsequent layers in a feed-forward manner makes training easier and more accurate. This is motivated by three observations. First, there is an implicit deep supervision thanks to the short paths to all feature maps in the architecture. Second, direct connections between all layers help improving the flow of information and gradients throughout the entire network. Third, dense connections have a regularizing effect, which reduces the risk of over-fitting on tasks with smaller training sets.

\begin{figure}[h!]
\centerline{\includegraphics[width=1\linewidth]{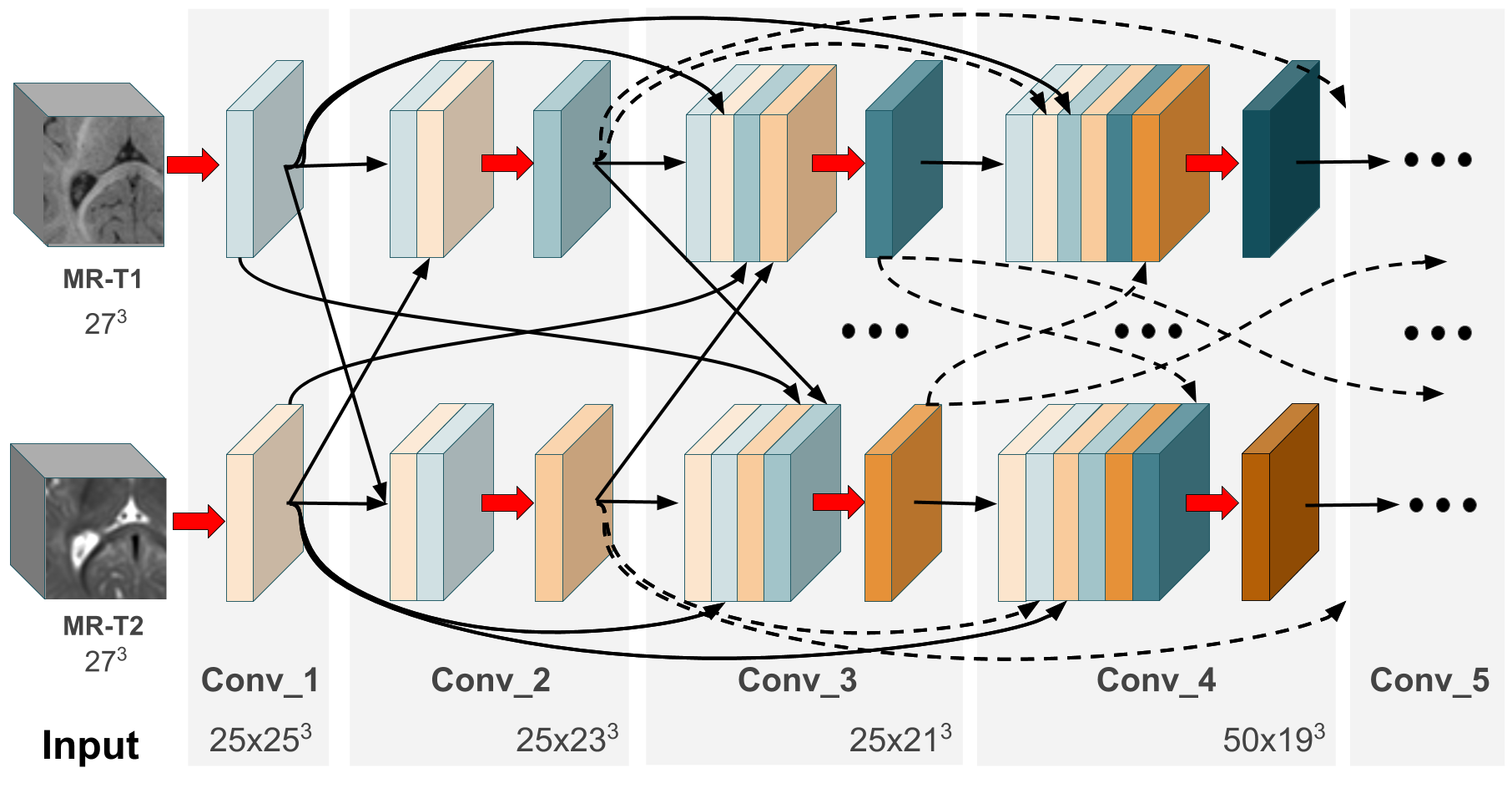}}
\caption{A section of the proposed \HyperDenseNet in the case of two image modalities. Each gray region represents a convolutional block. Red arrows correspond to convolutions and black arrows indicate dense connections between feature maps.}
\label{fig:net}
\end{figure}

Inspired by the recent success of densely-connected networks in medical image segmentation works \cite{li2017h,yu2017automatic,chenmri}, we propose a hyper-dense architecture for multi-modal image segmentation that extends the concept of dense connectivity to the multi-modal setting: each imaging modality has a path, and dense connections occur not only between layers within the same path, but also between layers across different paths (see Fig. \ref{fig:net} for an illustration).

Let $\xx_l$ be the output of the $l^{th}$ layer. In CNNs, this vector is typically obtained from the output of the previous layer $\xx_{l-1}$ by a mapping $H_l$ composed of a convolution followed by a non-linear activation function:
\begin{equation}
 \xx_l \ = \ H_l\big(\xx_{l-1}\big).
 \label{eq:layer_output}
\end{equation}
A densely-connected network concatenates all feature outputs in a feed-forward manner,
\begin{equation}
 \xx_l \ = \ H_l\big([\xx_{l-1}, \xx_{l-2}, \ldots, \xx_{0}]\big),
 \label{eq:layer_outputDense}
\end{equation}
where $[\ldots]$ denotes a concatenation operation. 

Pushing this idea further, HyperDenseNet introduces a more general connectivity definition, in which we link the outputs from layers in different streams, each associated with a different image modality. In the multi-modal setting, our hyper-dense connectivity yields a much more powerful feature representation than early/late fusion as the network learns the complex relationships between the modalities within and in-between all the levels of abstractions. For simplicity, let us consider the scenario of two image modalities, although extension to $N$ modalities is straightforward. Let $\xx_l^1$ and $\xx_l^2$ denote the outputs of the $l^{th}$ layer in streams 1 and 2, respectively. In general, the output of the $l^{th}$ layer in a stream $s$ can then be defined as follows:
\begin{equation}
 \xx_l^s \ = \ H_l^s\big([\xx_{l-1}^1, \xx_{l-1}^2, \xx_{l-2}^1, \xx_{l-2}^2, \ldots, \xx_{0}^1, \xx_{0}^2]\big).
 \label{eq:layer_HyperDense}
\end{equation}

Shuffling and interleaving feature map elements in a CNN was recently found to enhance the efficiency and performance, while serving as a strong regularizer \cite{zhang2017interleaved,chen2017regularization,zhang2017shufflenet}. This is motivated by the fact that intermediate CNN layers perform deterministic transformations to improve the performance, however, relevant information might be lost during these operations \cite{alain2016understanding}. To overcome this issue, it is therefore beneficial for intermediate layers to offer a variety of information exchange while preserving the aforementioned deterministic functions. Motivated by this principle, we thus concatenate feature maps in a different order for each branch and layer:
\begin{equation}
  \xx_l^s \ = \ H_l^s\big(\pi_l^s([\xx_{l-1}^1, \xx_{l-1}^2, \xx_{l-2}^1, \xx_{l-2}^2, \ldots, \xx_{0}^1, \xx_{0}^2])\big),
\end{equation}
with $\pi_l^s$ being a function that permutes the feature maps given as input. For instance, in the case of two image modalities, we could have: 
\begin{equation*}
\begin{split}
  \xx_l^1 & \ = \ 
    H_l^1\big([\xx_{l-1}^1, \xx_{l-1}^2, \xx_{l-2}^1, \xx_{l-2}^2, \ldots, \xx_{0}^1, \xx_{0}^2]\big) \\ 
  \xx_l^2 & \ = \ 
    H_l^2\big([\xx_{l-1}^2, \xx_{l-1}^1, \xx_{l-2}^2, \xx_{l-2}^1, \ldots, \xx_{0}^2, \xx_{0}^1])\big.
  \end{split}
\end{equation*}

Figure \ref{fig:net} shows a section of the proposed architecture, where each gray region represents a convolutional block. For simplicity, we assume that the red arrows indicate convolution operations only, whereas the black arrows represent the direct connections between feature maps from different layers, within and in-between the different streams. Thus, the input of each convolutional block (maps before the red arrow) is the concatenation of the outputs (maps after the red arrow) of all the preceding layers from both paths. 

\subsection{Baselines}
\label{ssec:Baselines}

\begin{figure*}[ht!]
\mbox{
 \shortstack{
 \includegraphics[width=0.3\linewidth]{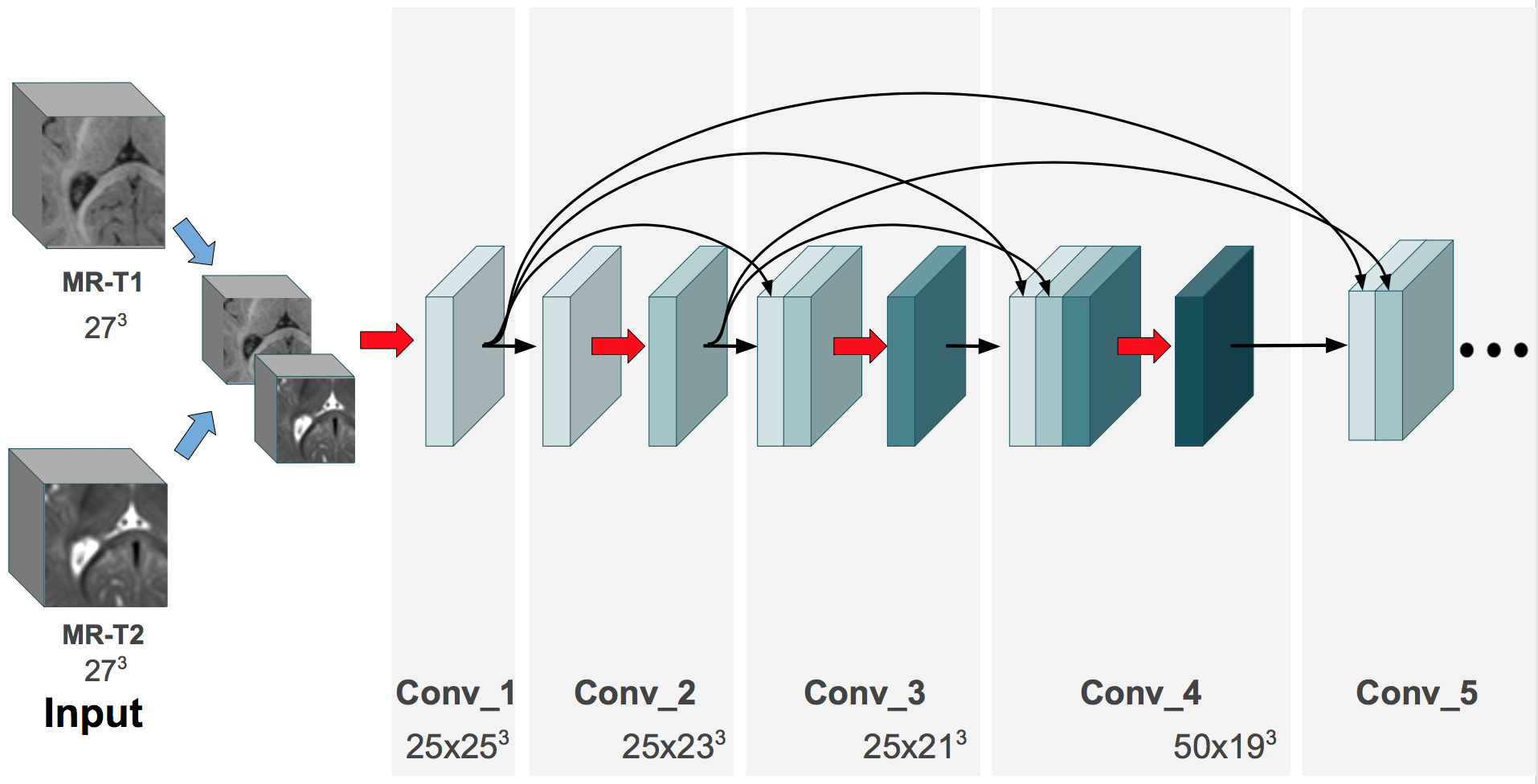}\\
 Single Dense Path 
 }
 \hspace{2.25 mm}
 \shortstack{
 \includegraphics[width=0.3\linewidth]{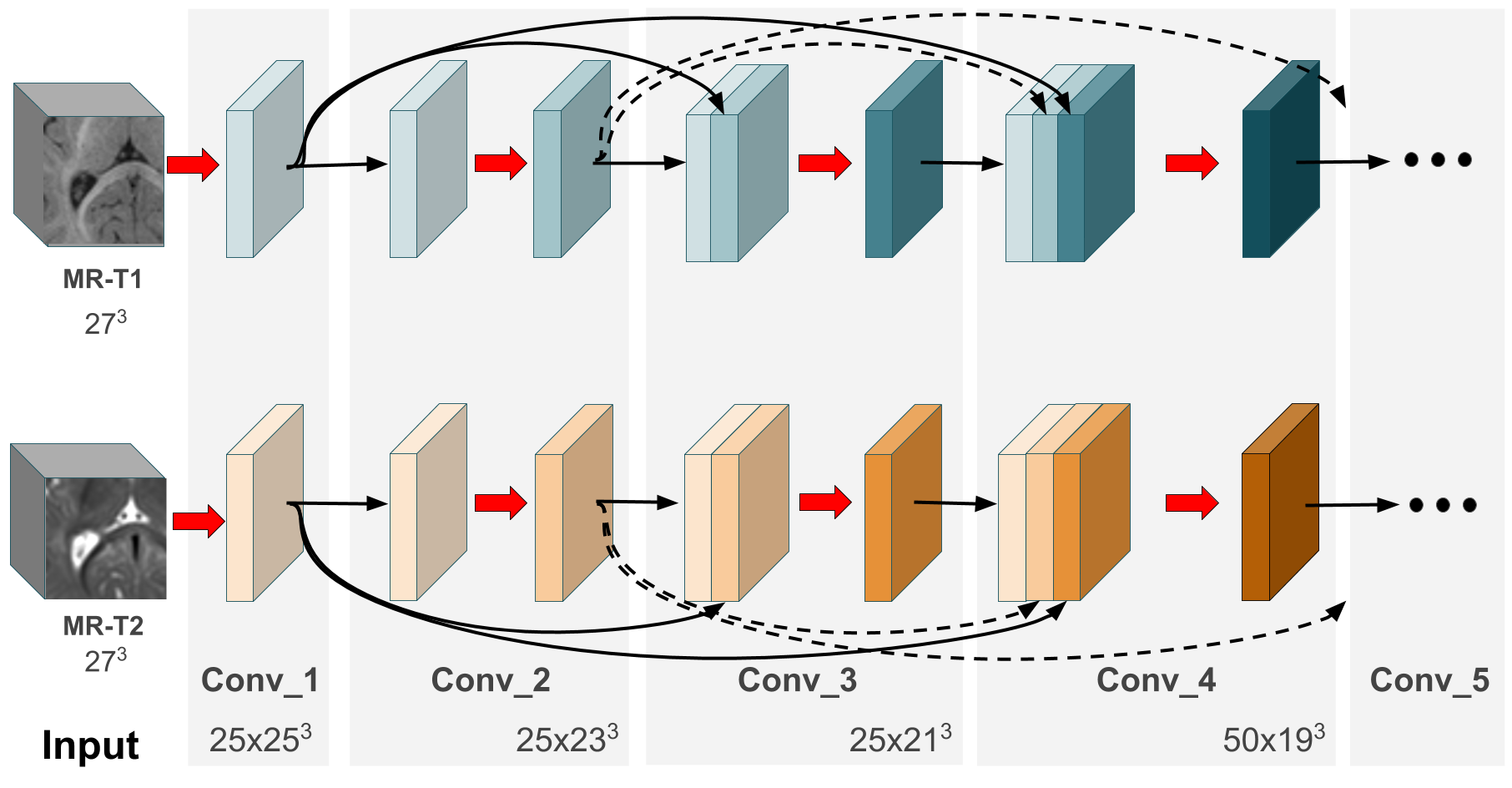}\\
 Dual Dense Path
 }
  \hspace{2.25 mm}
 \shortstack{
 \includegraphics[width=0.3\linewidth]{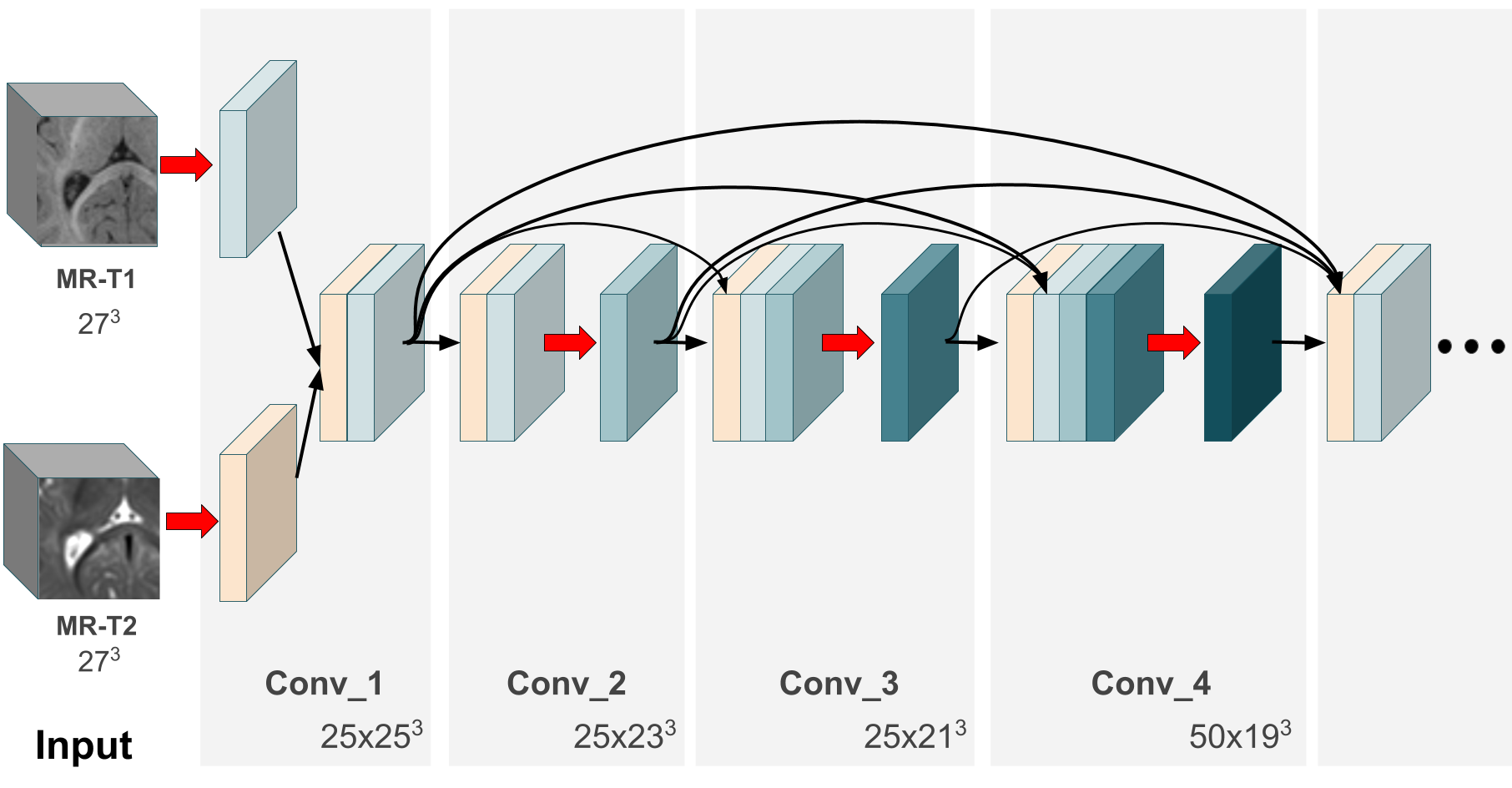}\\
 Disentangled modalities with early fusion
 }
 
 }
 \caption{Section of baseline architectures: single-path dense (\textit{left}), dual-path dense (\textit{middle}) with disentangled modalities and disentangled modalities with early fusion in a single path (\textit{right}). While both modalities are concatenated at the input of the network in the first case, each modality is analyzed independently in the second architecture with the features being fused at the end of the streams. Each gray region represents a convolutional block. Red arrows correspond to convolutions and black arrows indicate dense connections between feature maps. Dense connections are propagated through the entire network.}
\label{fig:Baselines}
\end{figure*}

To investigate thoroughly the impact of hyper-dense connections between different streams in multi-modal image segmentation, several baselines were considered. First, we extended the semi-dense architecture proposed in \cite{dolz2017Ensemble} to a fully-dense one, by connecting the output of each convolutional layer to all subsequent layers. In this network, we follow an early-fusion strategy, in which MRI T1 and T2 are integrated at the input of the CNN and processed jointly along a single path (Fig. \ref{fig:Baselines}, \textit{left}). The connectivity setting of this model corresponds to Eq. (\ref{eq:layer_outputDense}). Second, instead of merging both modalities at the input of the network, we considered a late-fusion strategy, where each modality is processed independently in different streams and learned features are fused before the first fully connected layer (Fig. \ref{fig:Baselines}, \textit{middle}). In this model, the dense connections are included within each path, assuming the connectivity definition of Eq. (\ref{eq:layer_outputDense}) for each stream. 

As last baseline, we used an early fusion model which combines features from different streams \emph{after the first convolutional layer} (Fig. \ref{fig:Baselines}, \textit{right}). Since this non-linear combination of features is re-used in all subsequent layers, the resulting network is similar to our hyper-dense model of Eq. (\ref{eq:layer_HyperDense}). However, there are two important differences. First, each stream in our model processes its input differently, as shown by the stream-indexed function $H_l^s$ in Eq. (\ref{eq:layer_HyperDense}). Also, as described above, each stream performs a different shuffling of inputs, which can enhance robustness to the model and mitigate the risk of overfitting. Our experiments in Section \ref{sec:experiments} demonstrate empirically the advantages of our model compared to this baseline. 

\subsection{Network architecture}

To have a large receptive field, FCNNs typically use full images as input. The number of parameters is then limited via pooling/unpooling layers. A problem with this approach is the loss of resolution from repeated down-sampling operations. In the proposed method, we follow the strategy in \cite{kamnitsas2017efficient}, where sub-volumes are used as input, avoiding pooling layers. While sub-volumes of size \vold{27} are considered for training, we used \vold{35} non-overlapping sub-volumes during inference, as in \cite{kamnitsas2017efficient,DolzNeuro2017}. This strategy offers two considerable benefits. First, it reduces the memory requirements of our network, thereby removing the need for spatial pooling. More importantly, it substantially increases the number of training examples and, therefore, does not need data augmentation.

\begin{table}[ht!]
\centering
\scriptsize
\caption{The layers used in the baselines and the proposed architecture and the corresponding values with an input of size \vold{27}. In the case of multi-modal images, the convolutional layers (conv\_x) are present in any network path. All the convolutional layers have a stride of one pixel.}
\renewcommand{\arraystretch}{1.1}
\begin{tabular}{lcccc}
\toprule
 & \multicolumn{1}{l}{\textbf{Conv. kernel}} & \multicolumn{1}{l}{\textbf{\# kernels}} & \multicolumn{1}{l}{\textbf{Output Size}} & \multicolumn{1}{l}{\textbf{Dropout}} \\ 
\midrule\midrule
\textbf{conv\_1} & 3$\times$3$\times$3 & 25 & \vold{25} & No \\ 
\textbf{conv\_2} & 3$\times$3$\times$3 & 25 & \vold{23} & No \\ 
\textbf{conv\_3} & 3$\times$3$\times$3 & 25 & \vold{21} & No \\ 
\textbf{conv\_4} & 3$\times$3$\times$3 & 50 & \vold{19} & No \\ 
\textbf{conv\_5} & 3$\times$3$\times$3 & 50 & \vold{17} & No \\ 
\textbf{conv\_6} & 3$\times$3$\times$3 & 50 & \vold{15} & No \\ 
\textbf{conv\_7} & 3$\times$3$\times$3 & 75 & \vold{13} & No \\ 
\textbf{conv\_8} & 3$\times$3$\times$3 & 75 & \vold{11} & No \\ 
\textbf{conv\_9} & 3$\times$3$\times$3 & 75 & \vold{9} & No \\ 
\textbf{fully\_conv\_1} & 1$\times$1$\times$1 & 400 & \vold{9} & Yes \\ 
\textbf{fully\_conv\_2} & 1$\times$1$\times$1 & 200 & \vold{9} & Yes \\ 
\textbf{fully\_conv\_3} & 1$\times$1$\times$1 & 150 & \vold{9} & Yes \\ 
\textbf{Classification} & 1$\times$1$\times$1 & 4 & \vold{9} & No \\ 
\bottomrule
\end{tabular}
\label{table:layers}
\end{table}

Table \ref{table:layers} summarizes the parameters of the baselines and the proposed HyperDenseNet. The network parameters are optimized via the RMSprop optimizer, using cross-entropy as cost function. Let $\vt$ denotes the network parameters (i.e., convolution weights, biases and $a_i$ from the parametric rectifier units), and $y^v_s$ the label of voxel $v$ in the $s$-th image segment. 
We optimize the following:
\begin{equation}
 J(\vt) \ = \ 
 -\frac{1}{S\!\cdot\!V} \sum^{S}_{s=1} \sum^{V}_{v=1} \sum^{C}_{c=1} \delta(y^v_s = c) \cdot \log \, p^v_c(\xx_s),
\end{equation}
where $p^v_c(\xx_s)$ is the softmax output of the network for voxel $v$ and class $c$, when the input segment is $\xx_s$. 

To initialize the weights of the network, we adopted the strategy proposed in \cite{he2015delving}, which yields fast convergence for very deep architectures. In this strategy, a zero-mean Gaussian distribution of standard deviation $\sqrt{2/n_l}$ is used to initialize the weights in layer $l$, where $n_l$ denotes the number of connections to the units in that layer. Momentum was set to 0.6 and the initial learning rate to 0.001, being reduced by a factor of 2 after every 5 epochs (starting from epoch 10). 
The network was trained for 30 epochs, each composed of 20 subepochs. At each subepoch, a total of 1000 samples were randomly selected from the training images and processed in batches of size 5. 

\section{Experiments and results}\label{sec:experiments}

The proposed \HyperDenseNet architecture is evaluated on challenging multi-modal image segmentation tasks, using publicly available data from two challenges: infant brain tissue segmentation, iSEG \cite{iSEG}, and adult brain tissue segmentation, MRBrainS\footnote{\url{http://mrbrains13.isi.uu.nl}}. Quantitative evaluations and comparisons with state-of-the-art methods are reported for each of these applications. First, to evaluate the impact of dense connectivity on performance, we compared the proposed \HyperDenseNet to the baselines described in Section \ref{ssec:Baselines} on infant brain tissue segmentation. Then, our results, compiled by the iSEG challenge organizers on testing data, are compared to those from the other competing teams. Second, to juxtapose the performance of \HyperDenseNet to other segmentation networks under the same conditions, we provide a quantitative analysis of the results of current state-of-the-art segmentation networks for adult brain tissue segmentation. This includes comparison to the participants the MRBrainS challenge. Finally, in Section \ref{ssec:featuresReuse}, we report a comprehensive analysis of feature re-use.

\subsection{iSEG Challenge}
\label{ssec:iSEG}
The focus of this challenge was to compare (semi-) automatic stat-of-the-art algorithms for the segmentation of 6-month infant brain tissues in T1- and T2-weighted brain MRI scans. This challenge was carried out in conjunction with MICCAI 2017, with a total of 21 international teams participating in the first round \cite{iSEG}.

\subsubsection{Evaluation}
\label{ssec:evalmetrics}

The iSEG-2017 organizers used three metrics to evaluate the accuracy of the competing methods: Dice Similarity Coefficient (DSC) \cite{dice1945measures}, Modified Hausdorff distance (MHD), where the 95-\textit{th} percentile of all Euclidean distances is employed, and Average Surface Distance (ASD). The first measures the degree of overlap between the segmentation region and ground truth, whereas the other two evaluate boundary distances.

\subsubsection{Results}

\paragraph*{\textbf{Validation results}} Table \ref{table:results} reports the performance achieved by \HyperDenseNet and the baselines introduced in Section \ref{ssec:Baselines}, for CSF, GM and WM brain tissues. The results were generated by splitting the 10 available iSEG-2017 volumes into training, validation and testing sets containing 6, 1 and 3 volumes, respectively. To show that improvements do not come from the higher number of learned parameters in HyperDenseNet, we also investigated a widened version of all baselines, with a similar parameter size as HyperDenseNet. The number of learned parameters of all the tested models is given in Table \ref{table:params}. A more detailed description of the tested architectures can be found in Table \ref{table:kernels} of the Supplemental materials ('Supplementary materials are available in the supplementary files /multimedia tab.').

We observe that the late fusion of deeper-layer features in independent paths provides a clear improvement over the single-path version, with an increase on performance of nearly 5$\%$. Fusing the feature maps from independent paths after the first convolutional layer (i.e., Dual-Single) outperformed the other two baselines by 1-2$\%$, particularly for WM and GM, which are the most challenging structures to segment. Also, the results indicate that processing multi-modal data in separate paths, while allowing dense connectivity between all the paths, increases performance over early and late fusion, as well as over disentangled modalities with fusion performed after the first convolutional block. Another interesting finding is that increasing the number of learned parameters does not bring an important boost in performance. Indeed, in some tissues (e.g., CSF for Single path and Dual-Single path architectures), the performance slightly decreased when widening the architecture.

\begin{table}[ht!]
\centering
\footnotesize
\caption{Performance on the testing set, in terms of DSC, for the investigated baselines and the proposed architecture. The best performance is highlighted in bold.}
\begin{tabular}{llccc}
\toprule
& \textbf{Architectures}                                                                                         & \textbf{CSF} & \textbf{WM} & \textbf{GM} \\
\midrule \midrule
\multirow{4}{*}{\begin{tabular}[c]{@{}l@{}}No connectivity\\ between paths\end{tabular}} 
& Single Path   & 0.9014  & 0.8518  & 0.8370 \\
& Single Path$^*$  & 0.9010  & 0.8532  & 0.8401 \\
& Dual Path & 0.9482  & 0.9078  & 0.8875\\
& Dual Path$^*$  & 0.9503  & 0.9089 & 0.8872 \\
\midrule
\multirow{3}{*}{\begin{tabular}[c]{@{}l@{}}Connectivity\\ between paths\end{tabular}} 
& Dual-Single Path   & 0.9552  & 0.9142  & 0.9008 \\
& Dual-Single Path$^*$ & 0.9541  & 0.9159  & 0.9017 \\
& HyperDenseNet   & \textbf{0.9580}  & \textbf{0.9183}   & \textbf{0.9035} \\
\bottomrule\\[-6pt]
$^*$ Widened version.
\end{tabular}
\label{table:results}
\end{table}

\begin{figure}[ht!]
\mbox{ \small 
 \includegraphics[width=1\linewidth]{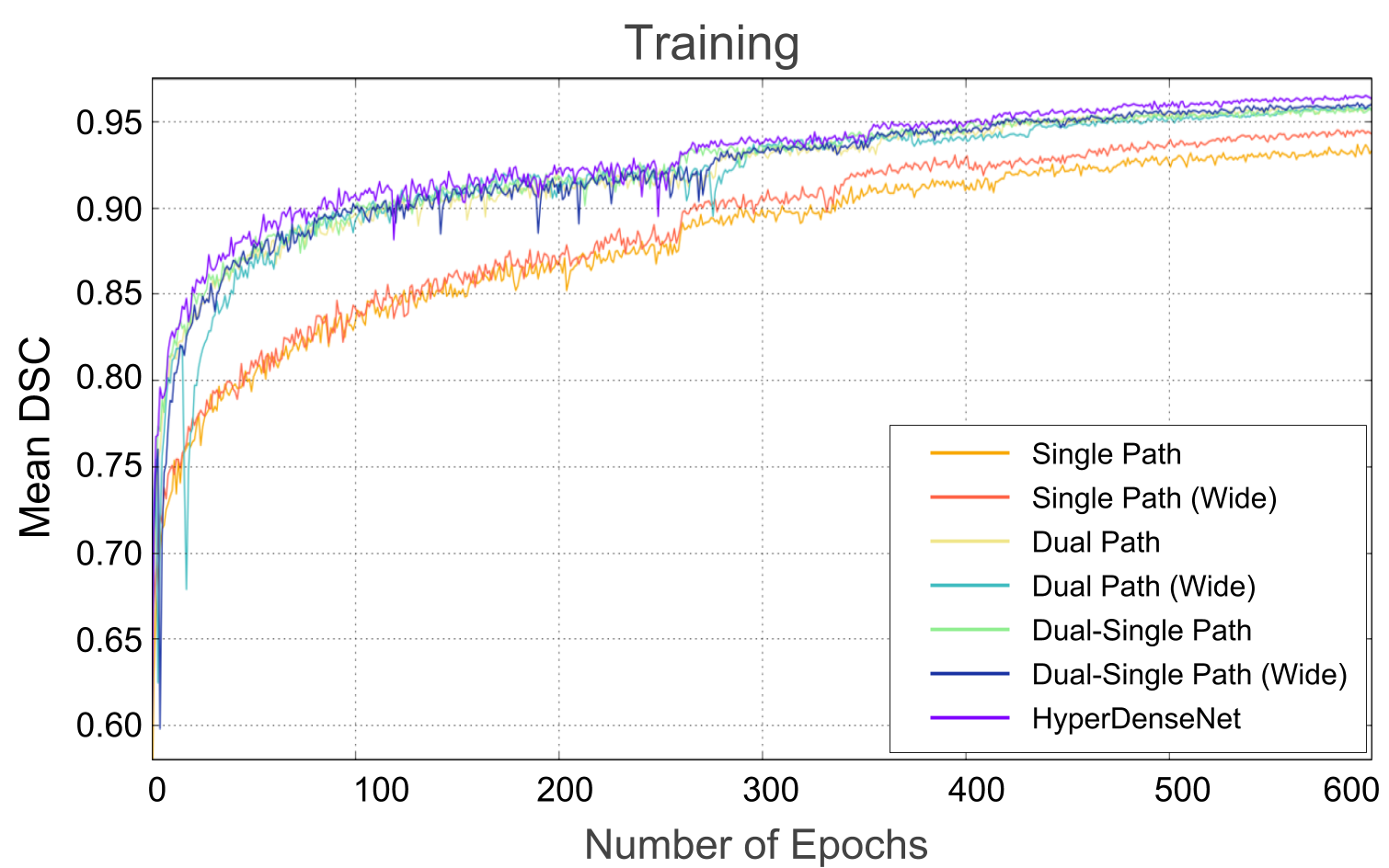} 
 }
 \vspace{-7mm}
 \caption{Training accuracy plots for the proposed architecture and the baselines on the iSeg-2017 challenge data. The first point of each curve corresponds to the end of the first training epoch.}
\label{fig:dsc_evol_train}
\end{figure}

\begin{figure}[ht!]
\mbox{ \small 
 \includegraphics[width=1\linewidth]{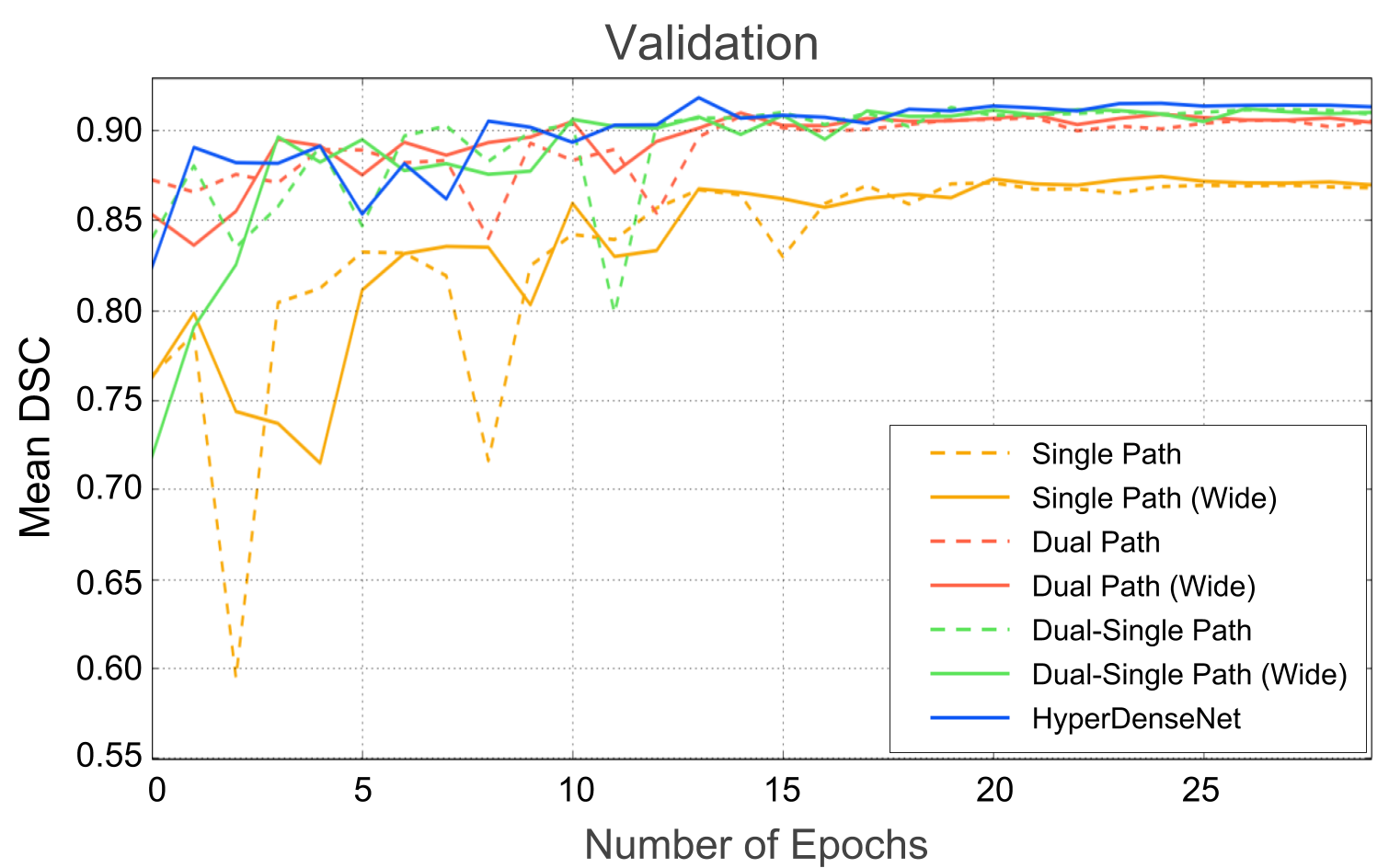} 
 }
 \vspace{-7mm}
 \caption{Validation accuracy plots for the proposed architecture and the baselines on the iSeg-2017 challenge data. The first point of each curve corresponds to the end of the first training epoch.}
\label{fig:dsc_evol}
\end{figure}

Figures \ref{fig:dsc_evol_train} and \ref{fig:dsc_evol} compare the training and validation accuracy between the baselines and \HyperDenseNet. In these figures, the mean DSC for the three brain tissues is evaluated during training (\textit{Top}) and validation (\textit{Bottom}). One can see that \HyperDenseNet outperforms baselines in both cases, achieving better results than architectures with a similar number of parameters. Performance improvements seen in Table \ref{table:results}, Fig. \ref{fig:dsc_evol_train} and Fig. \ref{fig:dsc_evol} might be due to two factors: the high number of direct connections between different layers, which facilitates back-propagation of the gradient to shallow layers, and the freedom of the network to explore more complex patterns thanks to the combination of several image modalities at any level of abstraction.



The computational efficiency of HyperDenseNet and baselines is compared in Table \ref{table:params}. As expected, inference times are proportional to the number of model parameters. While the lightest architecture needs around 45 seconds to segment a whole 3D brain, HyperDenseNet performs the same task in less than 2 minutes. This is acceptable from a clinical point of view.

\begin{table*}[ht!]
\centering
\footnotesize
\caption{Number of parameters (convolution, fully-connected and total) and inference times of the baselines and the proposed architecture. Widened versions of the baselines, which we denoted using superscript $^*$, are also included.}
\begin{tabular}{lcccc}
\toprule        
\multirow{2}{*}{\textbf{Architecture}} & \multicolumn{3}{c}{\textbf{Nb. of parameters}} & \multirow{2}{*}{\textbf{Time (sec)}} \\
\cmidrule(l{5pt}r{5pt}){2-4}
 & \textbf{Conv.} & \textbf{Fully-conn.} & \textbf{Total} & \\
\midrule \midrule
Single Path  &  2,380,050   & 290,600  & 2,670,650 & 43.67 ($\pm$8.37)\\
Single Path$^*$ & 9,518,850  & 470,600   & 9,989,450 & 101.63 ($\pm$12.65)\\
Dual Path   &  4,760,100  &  470,600  &  5,230,700   & 64.57 ($\pm$9.45)\\
Dual Path$^*$   & 9,381,960 &  614,600 &  9,996,560 & 104.31 ($\pm$11.65)\\
Dual-Single Path   & 2,666,760   & 300,600  & 2,968,200   & 47.33 ($\pm$8.74)\\
Dual-Single Path$^*$ & 9,518,850 & 470,600 & 9,989,450 & 103.64 ($\pm$13.61) \\
HyperDenseNet & 9,518,850  & 830,600  & 10,349,450 & 105.67 ($\pm$14.74) \\
\bottomrule\\[-6pt]
$^*$ Widened version.
\end{tabular}
\label{table:params}
\end{table*}

\begin{figure*}[ht!]
 \centering
 \includegraphics[width=0.8\linewidth]{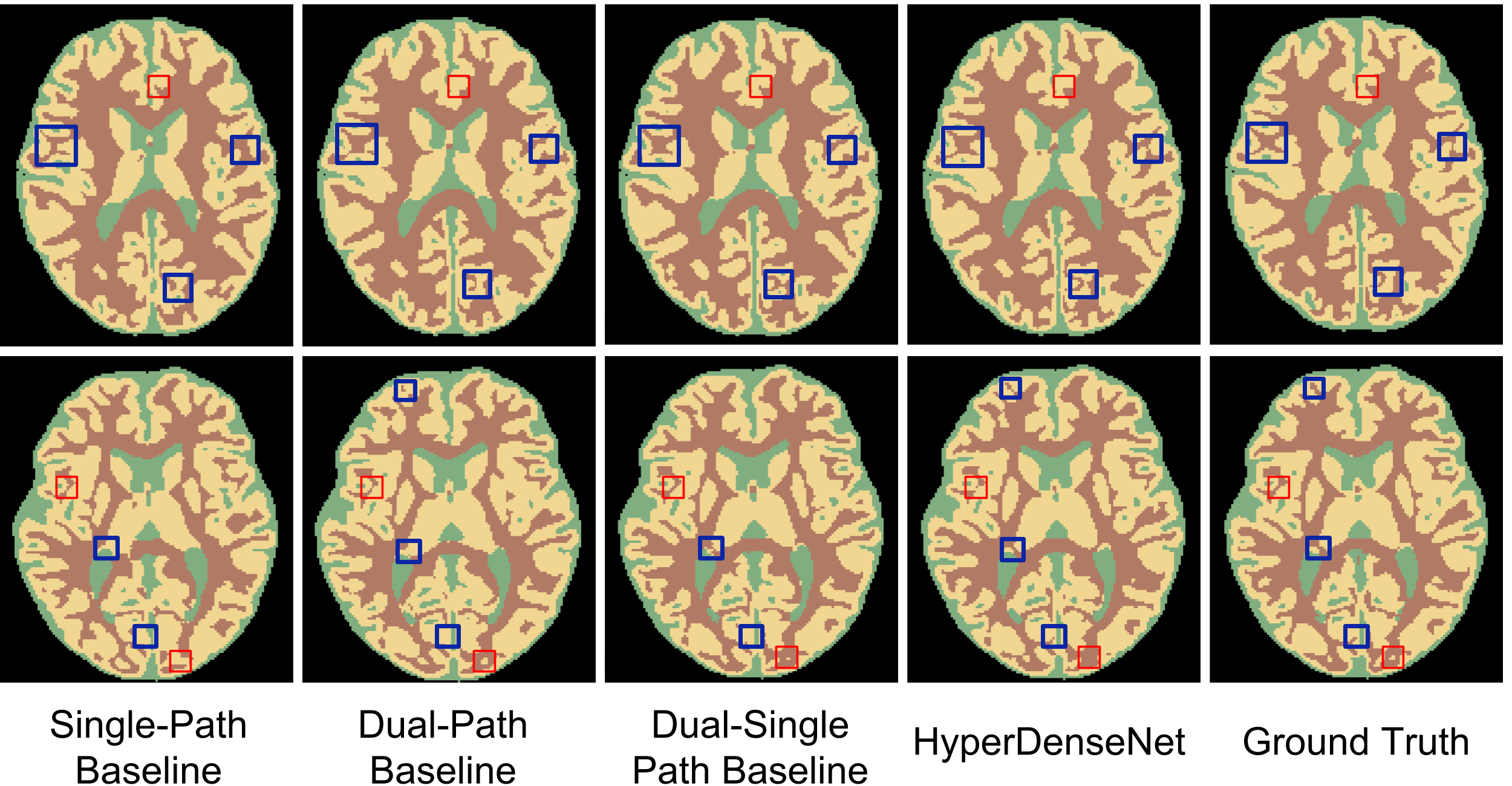} 
 \caption{Qualitative results of segmentation achieved by the baselines and \HyperDenseNet on two validation subjects (each row shows a different subject). The green squares indicate some spots, where \HyperDenseNet successfully reproduced the ground-truth whereas the baselines failed. Some regions where HyperDenseNet yielded incorrect segmentations are outlined in red.}
\label{fig:results}
\end{figure*}

Figure \ref{fig:results} depicts visual results for the subject used in validation. It can be seen that, in most cases, \HyperDenseNet typically recovers thin regions better than the baselines, which can explain the improvements observed for distance-based metrics. As confirmed in Table \ref{table:results}, this effect is most prominent in the boundaries between the gray and white matter. Furthermore, \HyperDenseNet produces fewer false positives for WM than the baselines, which tend to over-estimate the segmentation in this region.

\paragraph*{\textbf{Challenge results}} Table \ref{table:resultsChallenge} compares the segmentation accuracy of \HyperDenseNet to that of top-5 ranking methods in the first round of the iSEG Challenge, as well as to all the methods in the second round of submission. We observe that our network ranked among the top-3 methods in 6 out of 9 metrics, considering the results of the first and second rounds of submissions.

A noteworthy point is the general performance decrease of all the methods for the segmentation of GM and WM, with lower DSC and larger ASD values. This confirms that segmenting these tissues is more challenging due to the unclear boundaries between them. 

\begin{table*}[ht!]
\scriptsize
\centering
\caption{Results on the iSEG-2017 data for \HyperDenseNet and the methods ranked in the top-5 at the first round of submissions (in alphabetical order). The bold fonts highlight the best performances.
\textbf{Note}: The reported values were obtained from the challenge organizers at the time of submitting this manuscript, in February 2018. For an updated ranking, see the iSEG-2017 Challenge website for first (\url{http://iseg2017.web.unc.edu/rules/results/}) and second (\url{http://iseg2017.web.unc.edu/evaluation-on-the-second-round-submission/}) rounds of submission. The method referred to as LIVIA is a previous work from our team \cite{dolz2017Ensemble}.}
\label{table:resultsChallenge}
\begin{footnotesize}
\renewcommand{\arraystretch}{1.1}\setlength{\tabcolsep}{4.5pt}
\begin{tabular}{lccccccccccc}
\toprule
\multirow{2}{*}{\textbf{Method}} & \multicolumn{3}{c}{\textbf{CSF}} & & \multicolumn{3}{c}{\textbf{GM}} & & \multicolumn{3}{c}{\textbf{WM}} \\
\cmidrule(lr){2-4}\cmidrule(lr){6-8}\cmidrule(lr){10-12}
& \textbf{DSC} & \textbf{MHD} & \textbf{ASD} & & 
\multicolumn{1}{l}{\textbf{DSC}} & \textbf{MHD} & \textbf{ASD} & & 
\textbf{DSC} & \textbf{MHD} & \textbf{ASD} \\
\midrule 
\multicolumn{12}{l}{\textit{First round (Top 5)}}  \\ 

\midrule

Bern\_IPMI & 0.954 & 9.616 & 0.127 & & 0.916 & 6.455 & 0.341 & & 0.896 & 6.782 & 0.398 \\
LIVIA (ensemble) &  0.957 & 9.029 & 0.138 & & 0.919 & 6.415 & 0.338 & & 0.897 & 6.975 & 0.376\\
MSL\_SKKU & 0.958 & 9.072 & 0.116 & & 0.919 & 5.980 & 0.330 & & 0.901 & \textbf{6.444} & 0.391 \\
nic\_vicorob & 0.951 & 9.178 & 0.137 & & 0.910 & 7.647 & 0.367 & & 0.885 & 7.154 & 0.430 \\
TU/e IMAG/e & 0.947 & 9.426 & 0.150 & & 0.904 & 6.856 & 0.375 & & 0.890 & 6.908 & 0.433 \\
\midrule 
 \multicolumn{12}{l}{\textit{Second round (All methods)}} \\ 
 \midrule 
CatholicU & 0.916 & 10.970 & 0.241 & & 0.842 & 7.283 & 0.546 & & 0.819 & 8.239 & 0.675 \\
MSL\_SKKU & 0.958 & 9.112 & 0.116 & & 0.923 & 5.999 & 0.321 & & 0.904 & 6.618 & 0.375 \\
BCH\_CRL\_IMAGINE & \textbf{0.960} & \textbf{8.850} & \textbf{0.110} & & \textbf{0.926} & 9.557 & \textbf{0.311} & & \textbf{0.907} & 7.104 & \textbf{0.360} \\
\textbf{HyperDenseNet (Ours)}~ & 0.956 & 9.421 & 0.120 & & 0.920 & \textbf{5.752} & 0.329 & & 0.901 & 6.660 & 0.382\\

\bottomrule
\end{tabular}
\end{footnotesize}

\end{table*}

\subsection{MRBrainS Challenge}

The MRBrainS challenge was initially proposed in conjunction with MICCAI 2013. It focuses on adult brain tissue segmentation in the context of aging, based on three modalities: MRI T1, MRI T1 Inversion Recovery (IR) and MR-FLAIR. To this day, a total of 47 international teams have participated in this challenge.  

\subsubsection{Evaluation}

The organizers used three types of evaluation measures: a spatial overlap measure (DSC), a boundary distance measure (MHD) and a volumetric measure (the percentage of absolute volume difference).


\subsubsection{Architectures for comparison}

We compare \HyperDenseNet to three state-of-the-art networks for medical image segmentation. The first architecture is a 3D fully convolutional neural network with residual connections \cite{Residual}, which we denote as \textit{FCN\_Res3D}. The second one, referred to as \textit{UNet3D}, is a U-Net \cite{UNet3D} model with residual connections in the encoder and 3D volumes as input. Finally, our comparison also includes \textit{DeepMedic} \cite{kamnitsas2017efficient}, which showed an outstanding performance in brain lesion segmentation. The implementation details of these architectures are described in Supplemental materials (Supplementary materials are available in the supplementary files /multimedia tab).

\subsubsection{Results}


\paragraph*{\textbf{Validation results}} We performed a leave-one-out-cross-validation (LOOCV) on the 5 available MRBrainS datasets, using 4 subjects for training and one for validation. We trained and tested models three times, each time using a different subject for validation, and measured the average accuracy over these three folds. For this experiment, we used all three modalities (i.e., T1, T1 IR and FLAIR) for all competing methods.
In a second set of experiments, we assessed the impact of integrating multiple imaging modalities on the performance of \HyperDenseNet using all possible combinations of two modalities as input. 






\begin{table*}[ht!]
\centering
\caption{Comparison to several state-of-the-art 3D networks on the MRBrainS challenge.}
\label{table:resMRBrainS_DSC}
\renewcommand{\arraystretch}{1.1}
\begin{tabular}{lccc}
\toprule
\multirow{2}{*}{\textbf{Method}} & \multicolumn{3}{c}{\textbf{Mean DSC (std dev)}} \\ \cmidrule(lr){2-4}
 & \textbf{CSF} & \textbf{GM} & \textbf{WM} \\
\midrule\midrule
\textit{FCN\_Res3D} \cite{DLTK}~ (3-Modalities) & 0.7685 (0.0161)& 0.8163 (0.0222)& 0.8607 (0.0178)\\ 
\textit{UNet3D} \cite{UNet3D}~ (3-Modalities) & 0.8218 (0.0159) & 0.8432 (0.0241)& 0.8841 (0.0123)\\ 
DeepMedic\cite{kamnitsas2017efficient}~ (3-Modalities) & 0.8292 (0.0094) & 0.8522 (0.0193)& 0.8884 (0.0137)\\ 
\midrule
\HyperDenseNet~ (T1-FLAIR) & 0.8259 (0.0133) & 0.8620 (0.0260)& 0.8982 (0.0138) \\ 
\HyperDenseNet~ (T1\_IR-FLAIR) & 0.7991 (0.0181) & 0.8226 (0.0255)& 0.8654 (0.0087)\\ 
\HyperDenseNet~ (T1-T1\_IR) & 0.8191 (0.0297) & 0.8498 (0.0173) &0.8913 (0.0082) \\ 
\HyperDenseNet~ (3-Modalities) & \textbf{0.8485} (0.0078) & \textbf{0.8663} (0.0247) & \textbf{0.9016} (0.0109) \\ 
\bottomrule
\end{tabular}
\end{table*}

Table \ref{table:resMRBrainS_DSC} reports the mean DSC and standard-deviation values of tested models, with \textit{FCN\_Res3D} exhibiting the lowest mean DSC. This performance might be explained by the transpose convolutions in \textit{FCN\_Res3D}, which may cause voxel misclassification within small regions. Furthermore, the downsampling and upsampling operations in \textit{FCN\_Res3D} could make the feature maps in hidden layers sparser than the original inputs, causing a loss of image details. A strategy to avoid this problem is having skip connections as in \textit{UNet3D}, which propagate information at different levels of abstraction between the encoding and decoding paths. This can be be observed in the results, where \textit{UNet3D} clearly outperforms \textit{FCN\_Res3D} in all the metrics.

Moreover, \textit{DeepMedic} obtained better results than its competitors, yielding a performance close to the different two-modality configurations of \HyperDenseNet. The dual multiscale path is an important feature of \textit{DeepMedic} which gives the network a larger receptive field via two paths, one for the input image and the other processing a low-resolution version of the input. This, in addition to the removal of pooling operations in \textit{DeepMedic}, could explain the increase in performance with respect to \textit{FCN\_Res3D} and \textit{UNet3D}.

Comparing the different modality combinations, the two-modality versions of \HyperDenseNet yielded competitive performances, although there is a significant variability between the three configurations. 
Using only MRI T1 and FLAIR places \HyperDenseNet first for two DSC measures (GM and WM), and second for the remaining measure (CSF), even though competing methods used all three modalities. However, \HyperDenseNet with three modalities yields significantly better segmentations, with the highest mean DSC values for all three tissues. 

\paragraph*{\textbf{Challenge results}} The MRBrainS challenge organizers compiled the results and a ranking of 47 international teams\footnote{\url{http://mrbrains13.isi.uu.nl/results.php}}. In Table \ref{table:MRBrainSChallenge}, we report the results of the top-10 methods. We see that  \HyperDenseNet ranks first among competing methods, obtaining the highest DSC and HD for GM and WM. 
Interestingly, the BCH\_CRL\_IMAGINE and MSL\_SKKU teams participated in both iSEG and MRBrains2013 challenges. While these two networks outperformed HyperDenseNet in the iSEG challenge, the performance of our Model was noticeably superior in the MRBrains challenge, with HyperDenseNet ranked 1$^\textrm{st}$, MSL\_SKKU ranked 4$^\textrm{th}$ and BCH\_CRL\_IMAGINE ranked 18$^\textrm{th}$ (Ranking of February 2018). Considering the fact that three modalities are employed in MRBrains, unlike the two modalities used in iSEG, these results suggest that HyperDenseNet has stronger representation-learning power as the number of modalities increases.

\begin{table*}[ht!]
\centering
\footnotesize
\caption{Results of the MRBrainS challenge of different methods (DSC, HD (mm) and AVD). Only the top-10 methods are included in this table. \textbf{Note}: The reported values were obtained from the challenge organizers after submitting our results, in February 2018. For an updated ranking, see the MRBrainS Challenge website (\url{http://mrbrains13.isi.uu.nl/results.php}).}
\label{table:MRBrainSChallenge}
\renewcommand{\arraystretch}{1.1}\setlength{\tabcolsep}{4.5pt}
\begin{tabular}{lccccccccccccc}
\toprule
\multirow{2}{*}{\textbf{Method}} & \multicolumn{3}{c}{\textbf{GM}} & & \multicolumn{3}{c}{\textbf{WM}} & & \multicolumn{3}{c}{\textbf{CSF}} & & 
\multirow{2}{*}{\textbf{Sum}} \\
\cmidrule(lr){2-4}\cmidrule(lr){6-8}\cmidrule(lr){10-12}
 & \textbf{DSC} & \textbf{HD} & \textbf{AVD} & & 
 \textbf{DSC} & \textbf{HD} & \textbf{AVD}  & & 
 \textbf{DSC} & \textbf{HD} & \textbf{AVD} & & \\
\midrule\midrule
\textbf{HyperDenseNet (ours)} & \textbf{0.8633} & \textbf{1.34} & 6.19 & & \textbf{0.8946} & \textbf{1.78} & 6.03 & & 0.8342 & 2.26 & 7.31 & & \textbf{48} \\
VoxResNet \cite{chen2017voxresnet} + Auto-context~ & 0.8615 & 1.44 & 6.60 & & 0.8946 & 1.93 & 6.05 & & \textbf{0.8425} & 2.19 & 7.69 & & 54 \\
VoxResNet \cite{chen2017voxresnet} & 0.8612 & 1.47 & 6.42 & & 0.8939 & 1.93 & 5.84 & &  0.8396 & 2.28 & 7.44 & & 56 \\
MSL-SKKU & 0.8606 & 1.52 & 6.60 & & 0.8900 & 2.11 & \textbf{5.54} & & 0.8376 & 2.32 & 6.77 & & 61 \\
LRDE & 0.8603 & 1.44 & 6.05 & & 0.8929 & 1.86 & 5.83 & & 0.8244 & 2.28 & 9.03 & & 61 \\
MDGRU & 0.8540 & 1.54 & 6.09 & & 0.8898 & 2.02 & 7.69 & & 0.8413 & 2.17 & 7.44 & & 80 \\
PyraMiD-LSTM2 & 0.8489 & 1.67 & 6.35 & & 0.8853 & 2.07 & 5.93 & & 0.8305 & 2.30 & 7.17 & & 83 \\
3D-UNet \cite{UNet3D} & 0.8544 & 1.58 & 6.60 & & 0.8886 & 1.95 & 6.47 & & 0.8347 & 2.22 & 8.63 & & 84 \\
IDSIA \cite{stollenga2015parallel} & 0.8482 & 1.70 & 6.77 & & 0.8833 & 2.08 & 7.05 & & 0.8372 & \textbf{2.14} & 7.09 & & 100 \\
STH \cite{mahbod2018automatic} & 0.8477 & 1.71 & \textbf{6.02} & & 0.8845 & 2.34 & 7.67 & & 0.8277 & 2.31 & \textbf{6.73} & & 112 \\
\bottomrule
\end{tabular}
\end{table*}

\begin{figure}[ht!]
\centering
\includegraphics[width=1\linewidth]{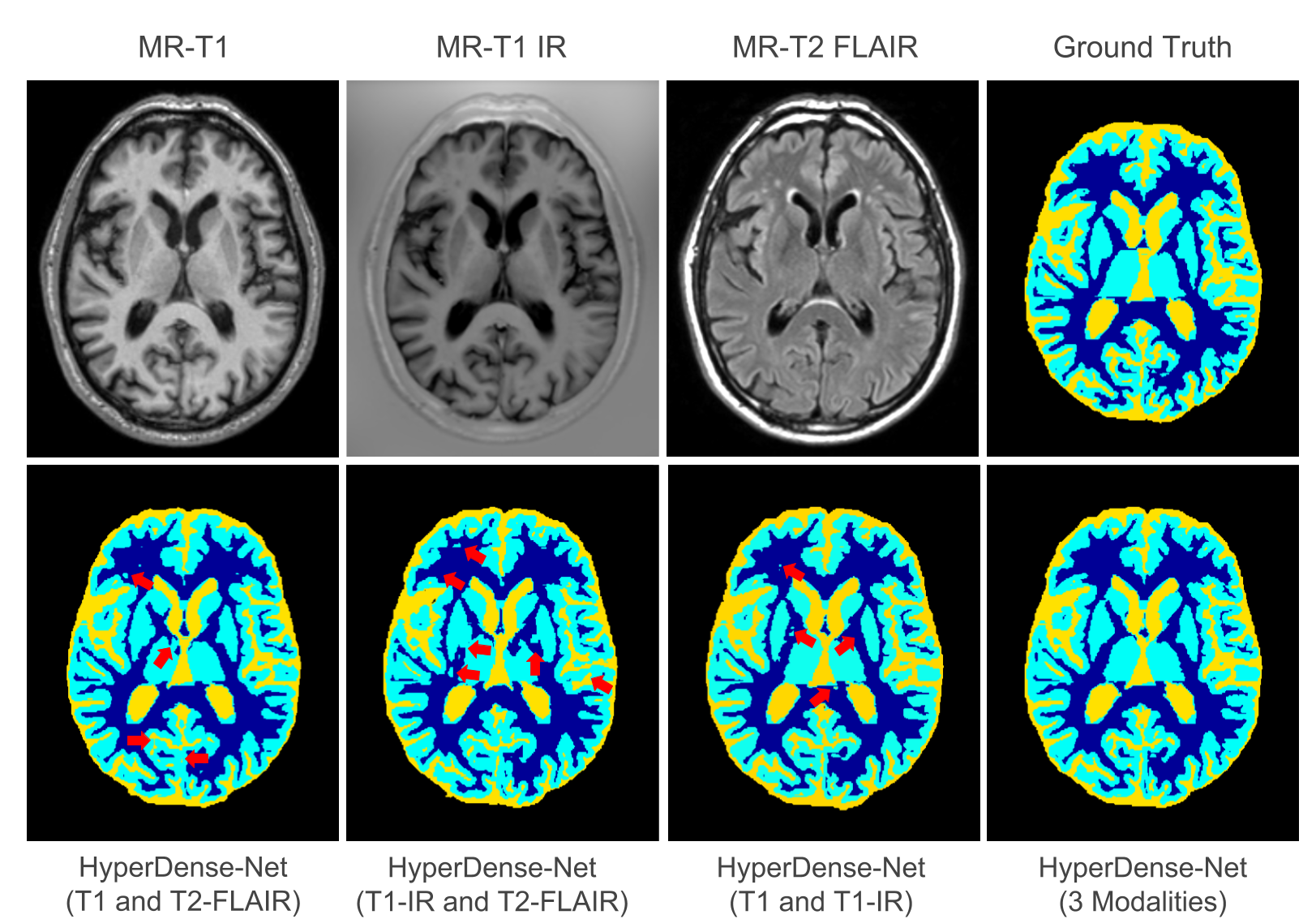}
\vspace{-7mm}
\caption{A typical example of the segmentations achieved by the proposed \HyperDenseNet in a validation subject (Subject 1 in the training set) for 2 and 3 modalities. 
The red arrows indicate some of the differences between the segmentations. For instance, one can see here that \HyperDenseNet with three modalities can handle thin regions 
better than its two-modality versions.}
\label{fig:MRBrainsSeg}
\end{figure}

A typical example of segmentation results is depicted in Fig. \ref{fig:MRBrainsSeg}. In these images, red arrows indicate regions where the two-modality versions of \HyperDenseNet fail in comparison to the three-modality version. As expected, most errors of these networks occur at the boundary between the GM and WM (see images in Fig. \ref{fig:isointense}, for example). Moreover, we observe that \HyperDenseNet using three modalities can handle thin regions better than its two-modality versions. 

\subsection{Analysis of features re-use}
\label{ssec:featuresReuse}

Dense connectivity enables each network layer to access feature maps from all its preceding layers, strengthening feature propagation and encouraging feature re-use. To investigate the degree at which features are used in the trained network, we computed, for each convolutional layer, the average \textit{L$_1$}-norm of connection weights to previous layers in any stream. This serves as a surrogate for the dependency of a given layer on its preceding layers. We normalized the values between 0 and 1 to facilitate visualization. 

\begin{figure}[ht!]
\centerline{\includegraphics[width=1\linewidth]{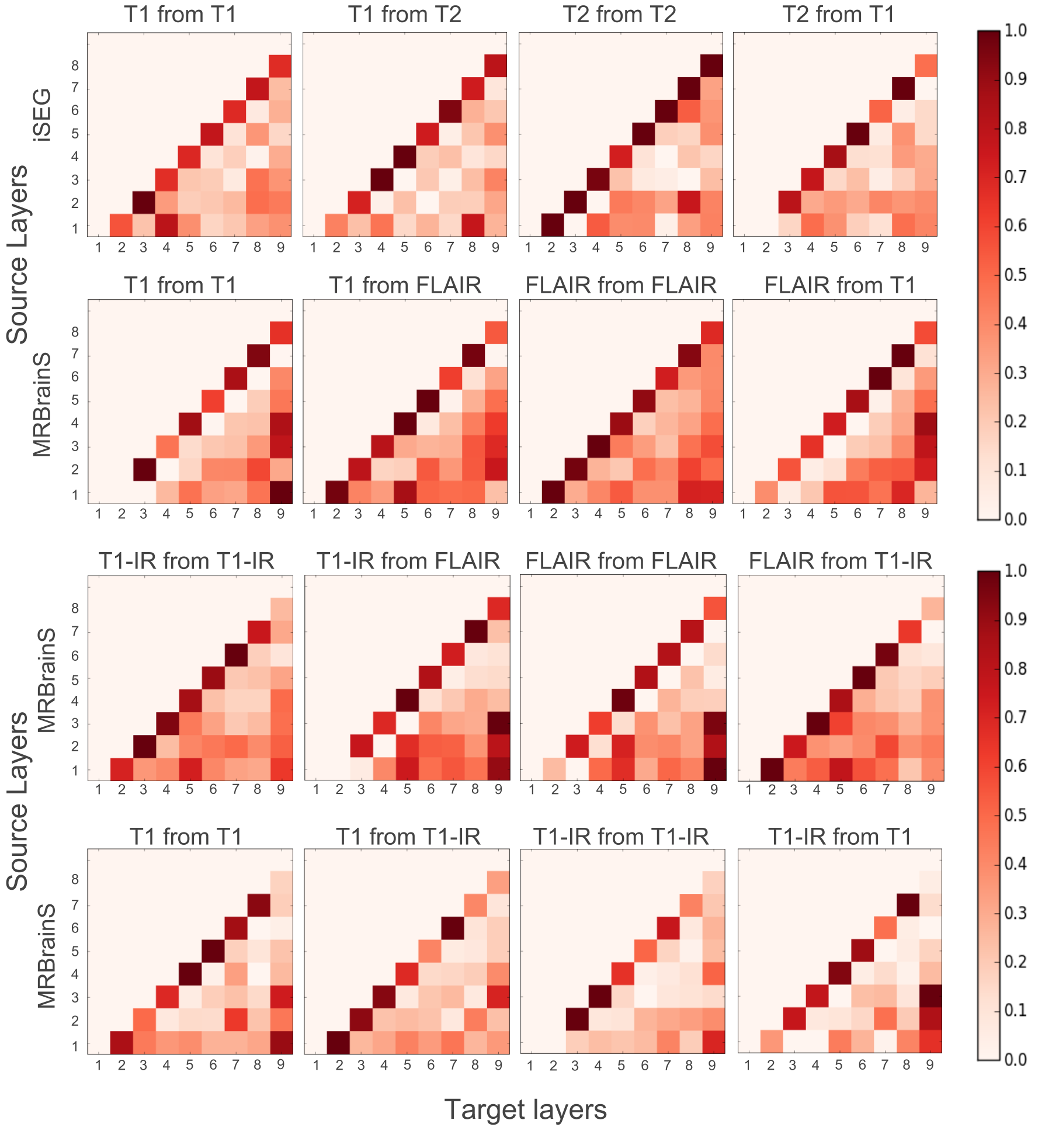}}
\vspace{-4mm}
\caption{Relative importance of connections in \HyperDenseNet trained on the iSEG (\textit{top}) and MRBrainS (\textit{from 2$^\textrm{nd}$ to 4$^\textrm{th}$ rows}) challenges with two modalities. The color at each location encodes the average L1 norm of weights connecting a convolutional-layer source to a convolutional-layer target. These values were normalized between 0 and 1 by accounting for all the values within each layer.}
\label{fig:weights}
\end{figure}

\begin{figure}[ht!]
\centerline{\includegraphics[width=0.8\linewidth]{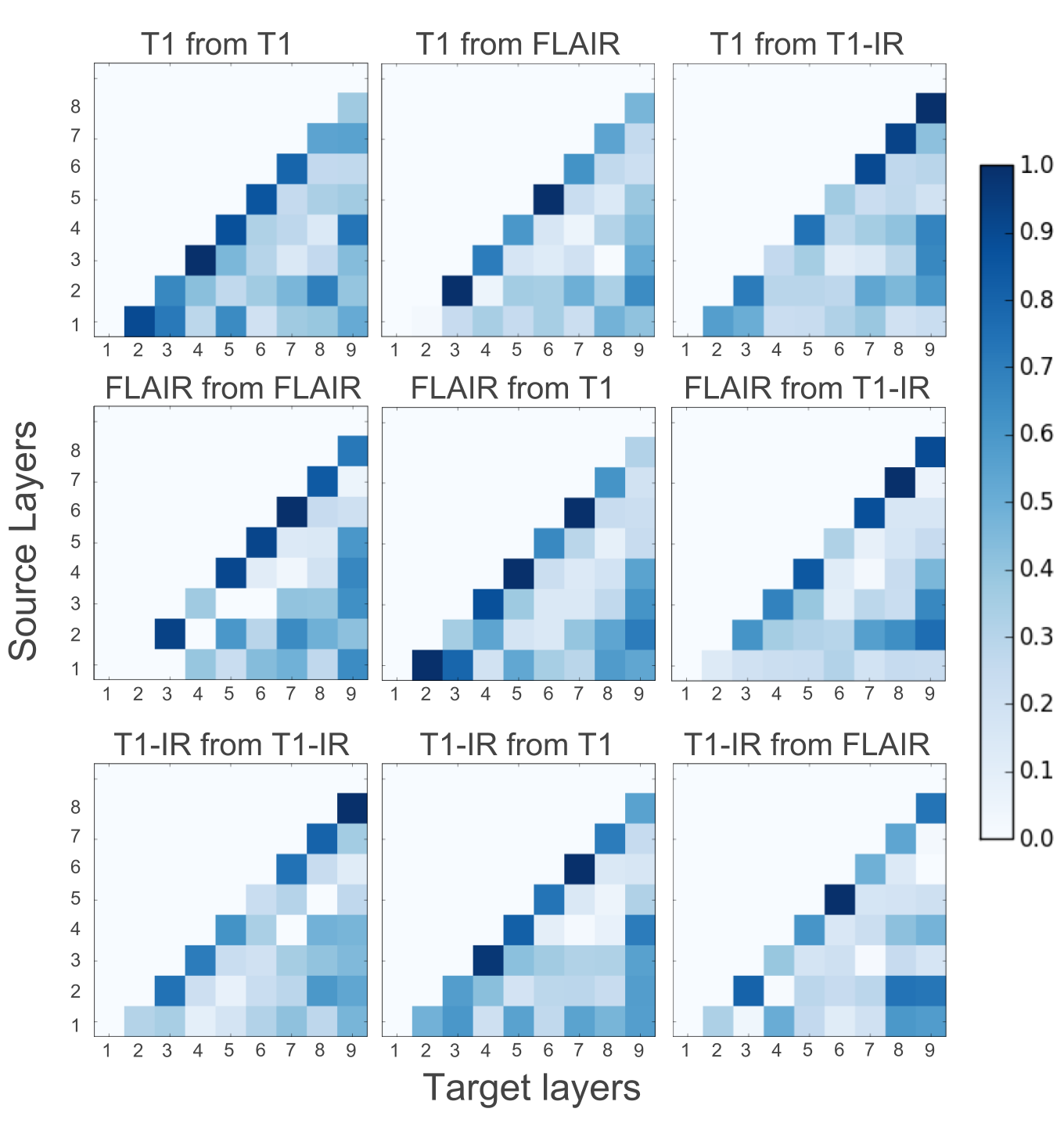}}
\vspace{-4mm}
\caption{Relative importance of connections in \HyperDenseNet trained on the MRBrainS challenge with three modalities (MRI T1, FLAIR and T1 IR). The color at each location encodes the average L1 norm of weights connecting a convolutional-layer source to a convolutional-layer target. These values were normalized between 0 and 1 by accounting for all the values within each layer.}
\label{fig:weights3mod}
\end{figure}

Figure \ref{fig:weights} depicts the weights of \HyperDenseNet trained with two modalities, for both iSEG and MRBrainS challenges. As the MRBrainS dataset contains three modalities, we have three different two-modality configurations. The average weights for the case of three modalities are shown in Fig. \ref{fig:weights3mod}. A dark square in these plots indicates that the target layer (\textit{on x-axis}) makes a strong use of the features produced by the source layer (\textit{on y-axis}). An important observation that one can make from both figures is that, in most cases, all layers spread the importance of the connections over many previous layers, not only within the same path, but also from the other streams. This indicates that shallow layer features are directly used by deeper layers from both paths, which confirms the usefulness of hyper-dense connections for information flow and learning complex relationships between modalities within different levels of abstractions. 

Considering challenge datasets separately, for \HyperDenseNet trained on iSEG (top row of Fig \ref{fig:weights}), immediate previous layers have typically higher impact on the connections from both paths. Furthermore, the connections having access to MRI T2 features typically have the strongest values, which may indicate that T2 is more discriminative than T1 in this particular situation. We can also observe some regions with high ($>$\,0.5) feature re-use patterns from shallow to deep layers. 
The same behaviour is seen for \HyperDenseNet trained on two modalities from the MRBrainS challenge, where immediate previous layers have a high impact on the connections within and in-between the paths. The re-use of low-level features by deeper layers is more evident than in the previous case. For example, in \HyperDenseNet trained with T1-IR and FLAIR, deep layers in the T1-IR path make a strong use of features extracted in shallower layers of the same path, as well as in the path corresponding to FLAIR. 
This strong re-use of early features from both paths occurred across all tested configurations. The same pattern is observed when using three modalities (Fig \ref{fig:weights3mod}), with a strong re-use of shallow features from the network's last layers. This reflects the importance of giving deep layers access to early-extracted features. Additionally, it suggests that learning how and where to fuse information from multiple sources is more effective than combining these sources in early or late stages.




\section{Conclusion}


This study investigated a hyper-densely connected 3D fully CNN, \HyperDenseNet, with applications to brain tissue segmentation in multi-modal MRI. Our model leverages dense connectivity beyond recent works \cite{li2017h,yu2017automatic,chenmri}, exploiting the concept in multi-path architectures. Unlike these works, dense connections occur not only within the stream of individual modalities, but also across differents streams. This give the network total freedom to explore complex combinations between features of different modalities, within and in-between all levels of abstraction. We reported a comprehensive evaluation using the benchmarks of two highly competitive challenges, iSEG-2017 for 6-month infant brain 
segmentation and MRBrainS for adult data, and showed state-of-the-art performances of \HyperDenseNet on both datasets. Our experiments provided new insights on the inclusion of short-cut connections in deep neural networks for segmentating medical images, particularly in multi-modal scenarios. In summary, this work demonstrated the potential of \HyperDenseNet to tackle challenging medical image segmentation problems involving multi-modal volumetric data. 

\ifCLASSOPTIONcompsoc
 \section*{Acknowledgments}
\else
 \section*{Acknowledgment}
\fi

This work is supported by the National Science and Engineering Research Council of Canada (NSERC), discovery grant program, and by the ETS Research Chair on Artificial Intelligence in Medical Imaging. The authors would like to thank both iSEG and MRBrainS organizers for providing data benchmarks and evaluations.

\ifCLASSOPTIONcaptionsoff
 \newpage
\fi



\bibliographystyle{IEEEtran}
\bibliography{2018_HyperDenseNet_Extension}

%



%




\clearpage

\setcounter{page}{1}
\begin{center}
\textbf{\large Supplemental Materials}
\end{center}

\subsection*{Datasets}
\subsubsection*{iSEG} 

Images were acquired at the UNC-Chapel Hill on a Siemens head-only 3T scanner with a circular polarized head coil, and were randomly chosen from the pilot study of the Baby Connectome Project (BCP)\footnote{http://babyconnectomeproject.org}. During scan, infants were asleep, unsedated and fitted with ear protection, with the head secured in a vacuum-fixation device. T1-weighted images were acquired with 144 sagittal slices using the following parameters: TR/TE = 1900/4.38 ms, flip angle = 7$^\circ$ and resolution = 1$\times$1$\times$1 mm$^3$. Likewise, T2-weighted images were obtained with 64 axial slices, TR/TE = 7380/119 ms, flip angle = 150$^\circ$ and resolution =1.25$\times$1.25$\times$1.95 mm$^3$. T2 images were linearly aligned onto their corresponding T1 images. All the images were resampled into an isotropic 1$\times$1$\times$1 mm$^3$ resolution. Standard image pre-processing steps were then applied using in-house tools, including skull stripping, intensity inhomogeneity correction, and removal of the cerebellum and brain stem. For this application, $9$ subjects were employed for training and 1 for validation. To obtain manual annotations, the organizers used 24-month follow-up scans to generate an initial automatic segmentation for 6-month subjects by employing a publicly available software iBEAT \footnote{http://www.nitrc.org/projects/ibeat/}. Then, based on the initial automatic contours, an experienced neuroradiologist corrected manually the segmentation errors (based on both T1 and T2 images) and geometric defects via ITK-SNAP, with surface rendering.

\subsubsection*{MRBrainS}

20 subjects with a mean age of 71 $\pm$ 4 years (10 male, 10 female) were selected from an ongoing cohort study of older ($65-80$ years of age), functionally-independent individuals without a history of invalidating stroke or other brain diseases \cite{reijmer2013disruption}. To test the robustness of the segmentation algorithms in the context of aging-related pathology, the subjects were selected to have varying degrees of atrophy and white-matter lesions, and the scans with major artifacts were excluded. The following sequences were acquired and used for the evaluation framework: 3D T1 (TR: 7.9 ms, TE: 4.5 ms), T1-IR (TR: 4416 ms, TE: 15 ms, and TI: 400 ms) and T2- FLAIR (TR: 11000 ms, TE: 125 ms, and TI: 2800 ms). The sequences were aligned by rigid registration using Elastix \cite{klein2010elastix}, along with a bias correction performed using SPM8 \cite{penny2011statistical}. After the registration, the voxel size within all the provided sequences (T1, T1 IR, and T2 FLAIR) was 0.96$\times$0.96$\times$3.00 mm$^3$. Five subjects that were representative for the overall data (2 male, 3 female and varying degrees of atrophy and white-matter lesions) were selected for training. The remaining fifteen subjects were provided as testing data. While ground truth was provided for the 5 training subjects, manual segmentations were unknown for the testing data set. The following structures were segmented and were available for training: (a) cortical gray matter, (b) basal ganglia, (c) white matter, (d) white matter lesions, (e) peripheral cerebrospinal fluid, (f) lateral ventricles, (g) cerebellum and (h) brainstem. These structures can be merged into gray matter (a-b), white matter (c-d), and cerebrospinal fluid (e-f). The cerebellum and brainstem were excluded from the evaluation. Manual segmentations were drawn on the 3mm slice thickness scans by employing an in-house manual segmentation tool based on the contour segmentation objects tool in Mevislab\footnote{https://www.mevislab.de/}, starting with the inner most structures. While the outer border of the CSF was segmented using both T1 and T1 IR scans, the other regions were segmented on the T1 scan.

\subsection*{Performance metrics}

\subsubsection*{Dice similarity coefficient (DSC)}

Let $V_\mr{ref}$ and $V_\mr{auto}$ be, respectively, the reference and automatic segmentations of a given tissue class and for a given subject. The DSC for this subject is defined as
\begin{equation}
\label{eq:dice}
\mr{DSC}\big(V_\mr{ref}, V_\mr{auto} \big) \ = \ 
 \frac{2 \mid V_\mr{ref} \cap V_\mr{auto}\mid} {\mid V_\mr{ref}\mid +\mid V_\mr{auto}\mid}
\end{equation}
DSC values are within a $[0,1]$ range, 1 indicating perfect overlap and 0 corresponding to a total mismatch.

\subsubsection*{Average volume distance (AVD)}

Using the same definitions for $V_\mr{auto}$ and $V_\mr{ref}$, AVD corresponds to
\begin{equation}
\label{eq:avd}
\mr{AVD}\big(V_\mr{ref}, V_\mr{auto} \big) \ = \ 
 \frac{\mid V_\mr{ref} - V_\mr{auto}\mid} {V_\mr{ref}}\cdot 100
\end{equation}

\subsubsection*{Modified Hausdorff distance (MHD)}

Let $P_\mr{ref}$ and $P_\mr{auto}$ denote the sets of voxels within the reference and automatic segmentation boundary, respectively. MHD is given by
\begin{small}
\begin{equation}
\label{eq:hausd}
\mr{MHD}\big(P_\mr{ref}, P_\mr{auto} \big) \ = \ \max \Big\{ \max_{q \in P_\mr{ref}}d(q,P_\mr{auto}), \max_{q \in P_\mr{auto}}d(q,P_\mr{ref}) \Big\},
\end{equation}
\end{small}
where $d(q,P)$ is the point-to-set distance defined by: $d(q,P)=\min_{p \in P} \| q-p\|$, with $\|.\|$ denoting the Euclidean distance. Low MHD values indicate high boundary similarity. 

\subsubsection*{Average surface distance (ASD)}

Using the same notation as the Hausdorff distance above, the ASD corresponds to
\begin{small}
\begin{equation}
 \mr{ASD}\big(P_\mr{ref}, P_\mr{auto} \big) \ = \ 
 \frac{1}{|P_\mr{ref}|} \sum_{p \, \in \, P_\mr{ref}} 
 d(p, P_\mr{auto}),
\end{equation}
\end{small}
where $|.|$ denotes the cardinality of a set. In distance-based metrics, smaller values indicate higher proximity between two point sets and, thus, a better segmentation. 

\subsection*{Implementation details}

We extended our 3D FCNN architecture proposed in \cite{DolzNeuro2017}, which is based on Theano. The source code of this architecture is publicly available\footnote{https://github.com/josedolz/SemiDenseNet}. 
Training and testing was performed on a server equipped with a NVIDIA Tesla P100 GPU with 16 GB of RAM memory. Training \HyperDenseNet took around 70 min per epoch, and around 35 hours in total for the two-modality version. With three image modalities, training each epoch took nearly 3 hours. Inference on a whole 3D MR scan took on average from 70-80 to 250-270 seconds, for the two- and three-modality versions, respectively.

The number of kernels per layer in each of the baselines and the proposed network are detailed in Table \ref{table:kernels}. 

\begin{table}[h!]
\centering
\footnotesize
\caption{Number of kernels (in convolutional and fully-connected layers) of the baselines and the proposed architecture. The architecture with two paths have the same number of kernels in both paths for the same convolutional block.}
\begin{tabular}{lcc}
\toprule        
\textbf{Architecture} & \multicolumn{1}{c}{\textbf{Conv. kernels}} & \multicolumn{1}{c}{\textbf{Fully-conn. kernels}}  \\
\midrule \midrule
Single Path  &  [25,\,25,\,25,\,50,\,50,\,50,\,75,\,75,\,75]   & [400,\,200,\,150]  \\
Single Path$^*$   & [50,\,50,\,50,\,75,\,75,\,75,\,150,\,150,\,150]  & [400,\,200,\,150]   \\
Dual Path   &   [25,\,25,\,25,\,50,\,50,\,50,\,75,\,75,\,75]  &  [400,\,200,\,150] \\
Dual Path$^*$   & [40,\,40,\,40,\,70,\,70,\,70,\,100,\,100,\,100] &  [400,\,200,\,150]  \\
Dual-Single Path   & [25,\,25,\,25,\,50,\,50,\,50,\,75,\,75,\,75]   & [400,\,200,\,150] \\
Dual-Single Path$^*$ & [25,\,50,\,50,\,100,\,100,\,100,\,150,\,150,\,150] & [400,\,200,\,150]  \\
HyperDenseNet & [25,\,25,\,25,\,50,\,50,\,50,\,75,\,75,\,75]  & [400,\,200,\,150]    \\
\bottomrule\\[-6pt]
$^*$ Widened version.
\end{tabular}
\label{table:kernels}
\end{table}

\subsubsection*{FCN\_Res3D} 

The architecture of \textit{FCN\_Res3D} consists on 5 convolutional blocks with residual units on the encoder path, with 16, 64, 128, 256 and 512 kernels.
The decoding path contains 4 convolutional upsampling blocks, each composed of 4 kernels, one per class. 
At each residual block, batch normalization and a Leaky ReLU with a leakage value of 0.1 are employed before the convolution. Instead of including max-pooling operations 
to re-size the images, stride values of 2 $\times$ 2 $\times$ 2 are used in layers 2, 3 and 4. Volume size at the input of the network is 64 $\times$ 64 $\times$ 24. 
The implementation of this network is provided in \cite{DLTK} \footnote{https://github.com/DLTK/DLTK}. 

\subsubsection*{UNet3D} 

Although quite similar to \textit{FCN\_Res3D}, \textit{UNet3D} presents some differences, particularly in the decoding path. It contains 9 convolutional blocks 
in total, 4 in the encoding and 5 in the decoding path. The number of kernels in the encoding path are 32, 64, 128 and 256, with strides of 2 $\times$ 2 $\times$ 2 at layers 2, 3 and 4. 
In the decoding path, the number of kernels are 256, 128, 64, 32 and 4, from the first to the last layer. Furthermore, skip connections are added at the convolutional blocks of the same scale between the encoding and decoding paths. As in \textit{FCN\_Res3D}, batch normalization and a Leaky ReLU with a leakage value of 0.1 are employed before the convolution at each block. 
Volume size at the input of the network is also 64 $\times$ 64 $\times$ 24. The implementation is provided in \cite{DLTK}. 

\subsubsection*{DeepMedic} 

We used the default architecture of \textit{DeepMedic} in our experiments. This architecture includes two paths with 8 convolutional blocks: 30, 30, 40, 40, 40, 40, 50, 50 kernels of size 3$\times$3$\times$3. At the end of both paths, two fully connected convolutional layers with 150 1$\times$1$\times$1 filters each are added, before the last classification layer. The second path is used with a low-resolution version of the input at the first path, for a larger receptive field. The input patch size is 27$\times$27$\times$27 and 35$\times$35$\times$35 for training and segmentation, respectively. The official code \footnote{https://github.com/Kamnitsask/deepmedic} is employed to evaluate this architecture.

\end{document}